\documentclass{egpubl}
 
\JournalPaper         % uncomment for final version of Journal Paper
\usepackage[T1]{fontenc}
\usepackage{dfadobe}  
\usepackage[labelfont=bf]{caption}

\usepackage[normalem]{ulem}

\usepackage[dvipsnames]{xcolor} % Remove that in the end!
\newcommand{\hl}[1]{#1} %Uncomment to remove the blue color

\usepackage{booktabs,ragged2e}

\usepackage[flushleft]{threeparttable}

\usepackage{amssymb}

\usepackage{caption}
\usepackage[absolute,overlay]{textpos}
\usepackage{pbox}
\usepackage{everypage}

\newcommand{\stamp}[1][© 2024 The Eurographics Association and John Wiley \& Sons Ltd. This is the author's version of the article that has been published in Computer Graphics Forum. The final version of this record is available at: \href{https://doi.org/10.1111/cgf.15004}{\color{blue}10.1111/cgf.15004}]{%
\begin{textblock*}{140mm}(37mm,270mm)
\centering%
\small% 
\emph{#1}%
\end{textblock*}%
}

\AddEverypageHook{
  \stamp
}

\usepackage{booktabs}
\usepackage{multirow}

\usepackage{microtype}                 % use micro-typography (slightly more compact, better to read)
\PassOptionsToPackage{warn}{textcomp}  % to address font issues with \textrightarrow
\usepackage{textcomp}                  % use better special symbols
\usepackage{mathptmx}                  % use matching math font
\usepackage{times}                     % we use Times as the main font
\usepackage{tabu}                      % only used for the table example

\usepackage{amsmath}
\usepackage{algorithm}
\newcommand{\euler}{e}

\usepackage{cite}  % comment out for biblatex with backend=biber
\BibtexOrBiblatex
\electronicVersion
\PrintedOrElectronic
\ifpdf \usepackage[pdftex]{graphicx} \pdfcompresslevel=9
\else \usepackage[dvips]{graphicx} \fi
\graphicspath{{figures/}{pictures/}{images/}{./}} % where to search for the images

\usepackage[super]{nth}
\usepackage{csquotes}

\usepackage{cleveref}
\renewcommand{\autoref}{\Cref}

\usepackage{egweblnk}
\usepackage{longfbox}

\newfboxstyle{tight2}{padding=2pt,margin=0pt,baseline-skip=false}%

\definecolor{ColT1}{HTML}{1f77b4}
\definecolor{ColT2}{HTML}{ff7f0e}
\definecolor{ColT3}{HTML}{2ca02c}
\definecolor{ColT4}{HTML}{d62728}
\definecolor{ColT5}{HTML}{9467bd}
\definecolor{ColT6}{HTML}{8c564b}
\definecolor{ColT7}{HTML}{e377c2}
\definecolor{ColT8}{HTML}{7f7f7f}
\definecolor{ColT9}{HTML}{bcbd22}
\definecolor{ColT10}{HTML}{17becf}

\newlength{\boxh}
\newcommand{\circled}[1]{\raisebox{.0pt}{\Large \textcircled{\raisebox{-.0pt} {\small #1}}}}
\newcommand{\cit}[1]{``#1''}

\title[DeforestVis: Behavior Analysis of Machine Learning Models \\ with Surrogate Decision Stumps]%
      {DeforestVis: Behavior Analysis of Machine Learning Models \\ with Surrogate Decision Stumps}

\author[Chatzimparmpas et al.]
{\parbox{\textwidth}{\centering A. Chatzimparmpas\thanks{Corresponding author: angelos.chatzimparmpas@northwestern.edu}$^{1}$\orcid{0000-0002-9079-2376}, R. M. Martins$^{2}$\orcid{0000-0002-2901-935X}, A. C. Telea$^{3}$\orcid{0000-0003-0750-0502}, and A. Kerren$^{2,4}$\orcid{0000-0002-0519-2537}}
        \\
{\parbox{\textwidth}{\centering $^1$Department of Computer Science, Northwestern University, USA\\
\centering $^2$Department of Computer Science and Media Technology, Linnaeus University, Sweden\\
$^3$Department of Information and Computing Sciences, Utrecht University, The Netherlands\\
$^4$Department of Science and Technology, Link\"{o}ping University, Sweden}}
}

\begin{document}

\stamp

\maketitle

\begin{abstract}
   As the complexity of Machine Learning (ML) models increases and their application in different (and critical) domains grows, there is a strong demand for more interpretable and trustworthy ML. A direct, model-agnostic, way to interpret such models is to train surrogate models---such as rule sets and decision trees---that sufficiently approximate the original ones while being simpler and easier-to-explain. Yet, rule sets can become very lengthy, with many if-else statements, and decision tree depth grows rapidly when accurately emulating complex ML models. In such cases, both approaches can fail to meet their core goal---providing users with model interpretability. To tackle this, we propose DeforestVis, a visual analytics tool that offers summarization of the behavior of complex ML models by providing surrogate decision stumps (one-level decision trees) generated with the Adaptive Boosting (AdaBoost) technique. DeforestVis helps users to explore the complexity vs fidelity trade-off by incrementally generating more stumps, creating attribute-based explanations with weighted stumps to justify decision making, and analyzing the impact of rule overriding on training instance allocation between one or more stumps. An independent test set allows users to monitor the effectiveness of manual rule changes and form hypotheses based on case-by-case analyses. We show the applicability and usefulness of DeforestVis with two use cases and expert interviews with data analysts and model developers.

   \makeatletter

	\def\customclassification{\vskip 5.5pt\par\reset@font\rmfamily}
	\def\endcustomclassification{\relax}
	\makeatother


	\begin{customclassification}
	\end{customclassification}
\end{abstract} 

\section{Introduction} \label{sec:intro}
	In Machine Learning (ML), surrogate models (also called \emph{metamodels} or \emph{emulators}) are interpretable models trained to approximate the predictions of a typically a black box, more complex, so-called \emph{target model}\,\cite{Sobester2008Engineering,Molnar2020Interpretable}. Such simpler,  more \emph{transparent}, surrogate models (or surrogates in brief) are more easy to examine and interpret than the original more complex models. For example, a hard-to-understand Convolutional Neural Network (CNN) can be approximated with a surrogate decision tree trained on the CNN's output predictions\,\cite{Jia2020Visualizing}. Surrogates can also describe the behavior of target models by summarizing their predictions in terms of the features of a given data set and associating misclassifications in a test set with particular subgroups of training samples. Training a surrogate is model-agnostic---it uses no explicit knowledge of the target model, only access to its input data and predictions\,\cite{Molnar2020Interpretable}.

Surrogates offer both local and global explanations~\cite{Elshawi2019Interpretability,Molnar2020Interpretable}. Local models, such as LIME\,\cite{Ribeiro2016Why}, aim to explain and reason about specific predictions of a target model. Global surrogates, the focus of our work, are simpler models that try to explain the overall behavior and global predictions of target models\,\cite{Castro2019Surrogate}. While any ML model can be used as a surrogate, rule sets (or lists)\,\cite{Ming2019RuleMatrix,Frank1998Generating}, decision trees\,\cite{Castro2019Surrogate,Safavian1991A}, and Generalized Additive Models (GAMs)\,\cite{Hohman2019Gamut,Caruana2015Intelligible,Nori2019InterpretML} are three particularly effective strategies. GAMs provide interpretable model coefficients that capture nonlinear relationships between input features\,\cite{Hastie1986Generalized}. Yet, they do not show relationships between the input and output features. Tree-based rule extraction methods\,\cite{Sato2001Rule} are a universal and mature technique, especially since surrogate decision trees reflect well the human decision-making process. These solutions are typically used to explain deep learning or ensemble learning algorithms with state-of-the-art predictive performance but poor decision accountability\,\cite{Jia2020Visualizing}. Rule extraction is a generalizable method in theory because its surrogate modeling does not consider the inner workings of black boxes. Still, applying it directly to neural networks is impractical and may result in suboptimal performance due to the latter's inherent complexity and nonlinearity\,\cite{Hohman2019Visual,Chatzimparmpas2020The}. Surrogate decision trees generated from end-to-end CNNs are too vast to parse, even with a modified gradient-based approach resulting in hundreds of levels~\cite{Zhang2019Interpreting}. Yuan et al.\,\cite{Yuan2022Visual} found that domain experts only examined one or two features at a time via a system similar to ours designed for exploring hierarchical surrogate rulesets. They found that shorter, larger, rulesets outperformed lengthy, uninterpretable rules and fully-grown decision trees in terms of model interpretability. Given these, an open question is: \textbf{(RQ1) \emph{How to summarize the behavior of large-scale ML models while providing \textbf{detailed} but \textbf{ultra-compact} explanations on demand?}}

The challenge of building accurate surrogate models can be addressed from two perspectives. \emph{Top-down} approaches aim to fit the surrogate model to the whole target model to emulate its behavior\,\cite{Ming2019RuleMatrix}. \emph{Bottom-up} approaches aggregate the results of local surrogates tailored for individual data instances\,\cite{Collaris2022StrategyAtlas}. Hybrid approaches that first train a global surrogate model (top-down) and then allow reasoning about specific cases (bottom-up) can be better for expert users. Model explanation systems should be designed so that users can quickly understand how models predict and be able to tune their output\,\cite{Yuan2022Visual}. Such systems should also enable the exploration of more precise surrogate models. These, however, come with larger, more complex, decision trees which take more effort to understand and thus reduce the surrogate's added-value\,\cite{Yuan2022Visual}. Ideally, we want to retain as much information as possible while reducing the number of decision trees of the surrogate---a complexity-fidelity trade-off\,\cite{Velmurugan2021Evaluating}. Separately, to achieve good generalizability for the surrogate model, one needs to override a rule only after one examines the data distribution for each feature and understands the implications of their changes locally (for a specific decision) and globally (for all decisions). This leads to our second question: \textbf{(RQ2) \emph{How to effectively help users to inject their domain knowledge into machine-produced rules while monitoring the \textbf{local} and \textbf{global} impact of their adjustments?}}

We present \emph{\textsc{DeforestVis}} (see \autoref{fig:teaser}), a Visual Analytics (VA) tool for the exploratory analysis of \emph{decision stumps}---one-level decision trees~\cite{Iba1992Induction}. \textsc{DeforestVis} creates such stumps using the Adaptive Boosting (AdaBoost) method\,\cite{Schapire1999A}. Our tool allows users to trade off complexity of the visual explanation (number of stumps) \emph{vs} fidelity of the surrogate (accuracy score). \textsc{DeforestVis} summarizes the decision boundaries and the contribution of each feature to a separate test set. An in-depth analysis is possible by observing the influence of individual decision stumps. Also, users can visually inspect both the local and global impact of a change in a rule. In summary, our contributions are as follows:

\begin{itemize}
\item a visual analytic workflow that simplifies the behavior analysis of complex ML models via surrogate models;
\item an implementation of this workflow in a VA tool via multiple linked views for selecting accurate and simple surrogate models, summarizing the behavior of complex models while also explaining how the aggregated information was computed, and formulating what-if hypotheses when overriding particular rules extracted from decision stumps;
\item a proof-of-concept use case and a usage scenario with real-world healthcare data that highlight the efficiency and effectiveness of our approach in forming compact rule sets; and
\item the evaluation of our proposal via interviews with data analysts and model developers.
\end{itemize}

\noindent We organize the rest of this paper as follows. \autoref{sec:back} introduces the necessary background information for making this paper self-sustained. \autoref{sec:relwo} discusses related work on surrogate models for model interpretation and approaches for visualizing rules and decision trees. \autoref{sec:goals} describes the user goals, analytical tasks, and user types, of VA tools using surrogate models for behavior analysis of complex ML models. \autoref{sec:overview} presents our tool. \autoref{sec:case} describes a use case for examining alternative decisions and their combinatorial effect while manually adjusting a decision rule. It also shows the applicability and usefulness of \textsc{DeforestVis} with a real-world data set for a binary classification problem. \autoref{sec:eval} presents feedback obtained from expert interview sessions and reports the limitations identified by the experts. \autoref{sec:disc} reflects further on the targeted users and the limitations that may lead to improvements for our tool. Finally, \autoref{sec:con} concludes the paper.

\section{Background} \label{sec:back}
	We next briefly overview the algorithmic steps of the AdaBoost surrogate model used in \textsc{DeforestVis}, so as to introduce the reader to the utilized methods and measures. For more details, we refer to the original algorithm's paper from Freund et al.~\cite{Freund1999A}.

\noindent\textbf{Data set:} For a binary classification problem, let $T=\{x_i\}$, $1\leq i \leq N$, $T \subset \mathbb{R}^n$, be a training set of $n$-dimensional instances $x_i$, each having a label $y_i \in C$, where  $C =\{C_1,C_2\}$ for the binary case.

\noindent\textbf{Target model:} A (complex) model $f$ is trained on $T$ to  predict variables $y = f(x)$.

\noindent\textbf{Surrogate model:} AdaBoost successively fits many decision stumps (\emph{maximum depth} $=1$) to weighted training samples from $T$ and next tries to predict $y$.  Initially, all weights are uniformly set to $w_i = 1/N$. Should the initial stump's prediction prove inaccurate, the samples $x_i$ that were incorrectly predicted receive a higher weight $w_i$. \hl{Throughout the process}, we keep the constraint $\sum_i w_i = 1$ with $w_i \in [0,1]$. The above process iterates from $m = 1$ to $M$ and is regulated by a \emph{learning rate} hyperparameter. Adding decision stumps continues until reaching a user-defined limit set by a \emph{number of trees/estimators} hyperparameter.

\noindent\textbf{Choosing features and split points for a stump:} To build individual decision stumps, a node $v$ is selected for dividing data into two child nodes. For every $v$ in a stump, the best \emph{decision threshold} $p_v$ among the $n$ is chosen according to the \emph{Gini impurity}\,\cite{Genuer2010Variable} metric $GI(v) = \sum_{c \in C} P_{vc} (1 - P_{vc})$, where $P_{vc}$ is the likelihood of a specific classification outcome $c$ (in our case $C_1$ and $C_2$).

\noindent\textbf{Impact of weights:} Within the $m = 1$ to $M$ loop, a model $G_m(x)$ is fitted to the training set $T$ using the weights $w_i$, leading to the concept of \emph{weighted probability} ($W \times P$) that plays a central role in \textsc{DeforestVis}. Subsequently, AdaBoost uses this method to measure the training performance of the classifier given by $\alpha_m = \log((1 - \text{err}_m) / \text{err}_m)$, where $\text{err}_m$ is the misclassification rate for the training set at iteration $m$. 
Therefore, $\alpha_m$ gives the influence a particular stump will exert in the classification. When a stump classifies correctly, producing zero misclassifications, its error rate is 0 and its $\alpha$ value becomes a large positive number. Conversely, a stump with a 50\% correct classification will have an $\alpha$ value of 0. If a stump predominantly misclassifies, then $\alpha$ would become a large negative value.

\noindent\textbf{Final ensemble of weak learners:} After evaluating the  error values for every stump, the sample weights are revised by $w_i^{m+1} = w_i^{m} \cdot \euler^{\alpha_m}$. As described earlier, $\alpha$ is (1) positive when the prediction aligns with the actual output or (2) negative when there's a discrepancy. In case (1), the sample weight diminishes compared to its previous value due to the algorithm already performing well. In case (2), the sample weight rises to prevent a similar misclassification by the next stump. This mechanism ensures subsequent rastumps are influenced by their predecessors. Finally, AdaBoost's output aggregates all the stumps as $G(x) = \pm \bigl[ \sum_{m = 1}^M \alpha_m G_m(x) \bigr]$.

\noindent\textbf{Training multiple surrogate models:} In \textsc{DeforestVis}, we train multiple surrogate models with incrementally more stumps to explore the trade-off between \emph{complexity} (measured as the number of decision stumps) and \emph{fidelity} (measured as how accurately the surrogate model fits the target model's predictions)\,\cite{Castro2019Surrogate}). This leads to stumps that were found initially in the least complex surrogate model (\emph{unique stumps or rules}) to be repeated either in the subsequent surrogates (\emph{original stumps}) or the surrogate itself (\emph{duplicated stumps}). Following the algorithmic process explained above, \emph{duplicated stumps} use the same \emph{feature and decision threshold value} as the \emph{original stump}, but with different weights/voting power while possibly predicting the opposite class for the same segment. Hence, they aim to fix the error (i.e., smooth the effect) of an \emph{original stump}. In contrast, \emph{unique decision stumps} most likely have greater weight, thus contributing significantly to the final ensemble of stumps.

\section{Related Work} \label{sec:relwo}
	We next review related work on visualizing the internal structures and outputs of surrogate models (\autoref{sec:search}) and broader visualization techniques for decision trees and rules (\autoref{sec:ensem}). We also highlight our contributions in relation to existing work.

\subsection{Surrogate model visualization} \label{sec:search}
Recent work has used surrogate models to approximate the behavior of complex ML models locally~\cite{Agus2021RISSAD,Ribeiro2016Why,Ribeiro2018Anchors,Lundberg2017A,Thomas2021FacetRules,Yuan2022SUBPLEX,Eisemann2014A}, globally\,\cite{Cao2020DRIL,Yuan2022Visual,Castro2019Surrogate,Ming2019RuleMatrix}, or on all scales~\cite{Collaris2022StrategyAtlas,Jia2020Visualizing}; all of these examples provide visual exploration of such surrogates as well. Closer to our work, SuRE~\cite{Yuan2022Visual} uses hierarchical rules to describe the decision space of a given ML model and explore its results by an interactive hierarchical visualization of the extracted rules. However, when checking multiple intertwined rules in the form of if-else statements, participants evaluating SuRE almost always examines at most two conditions at a time -- a limitation we overcome with \textsc{DeforestVis} due to the simple nature of one-level decision trees. Another interesting finding is that these users analyzed thoroughly
the effect of predictions based on \emph{individual features}, thus matching well the main design concept of our proposed tool. Di Castro and Bertini~\cite{Castro2019Surrogate} use a single surrogate decision tree to replicate a classification model's prediction and visualize it to propose simple yet effective explanations for the original model. RuleMatrix~\cite{Ming2019RuleMatrix} uses a matrix design and Sankey diagram visualization for the content of a rule list showing how data flows through the list. The problem with the above two VA tools is that they use a flat tabular layout~\cite{Varu2022ARMatrix} which disregards the rules' hierarchical structure and the important feature-ordering information captured by the hierarchical structure of decision trees. In our approach, this is not a problem since AdaBoost produces one-level decision trees (stumps), and we sort stumps for the same feature on the \cit{importance} extracted directly from the AdaBoost algorithm. DRIL~\cite{Cao2020DRIL} presents a rule list for adjusting thresholds and examining relationships between rules and data. Our VA tool focuses on both the summarized rules and the decision stumps that serve as an extra explanation of how the aggregation of information occurs.

StrategyAtlas~\cite{Collaris2022StrategyAtlas}, a hybrid approach, aims to explain individual data instances by aggregating multiple local surrogates~\cite{Collaris2022StrategyAtlas}. The method employs well-known explanation techniques, such as LIME~\cite{Ribeiro2016Why} and SHAP~\cite{Lundberg2017A}, to obtain feature-vector contributions. Points with similar feature contributions are grouped together via dimensionality reduction. However, the final visualization produced by StrategyAtlas is a surrogate decision tree, which suffers from interpretability issues due to its if-else structure. Since we use shallow decision trees, our approach suffers far less from this problem. CNN2DT shows the data flow through the surrogate decision tree of a CNN by a collapsible tree~\cite{Jia2020Visualizing}. In contrast, our approach is not specific to a single model (e.g., CNNs) and can be adapted to suit a range of domains based on the data sets and the expertise of the expert user.

Among local surrogate approaches~\cite{Agus2021RISSAD,Ribeiro2016Why,Ribeiro2018Anchors,Lundberg2017A,Thomas2021FacetRules,Yuan2022SUBPLEX,Eisemann2014A}, SUBPLEX~\cite{Yuan2022SUBPLEX},  provides a visual explanation for interpreting sub-populations of local explanations. It uses clustering and projection visualization techniques to help users better understand these explanations. Yet, this approach trains a local surrogate, whereas ours aggregates the result with a global surrogate and provides explanations of individual data instances of interest to the user.

\subsection{Tree- and rule-based model visualization} \label{sec:ensem}
Many VA tools have been created to examine decision trees stemming from bagging~\cite{Zhao2019iForest,Neto2021Explainable,Eirich2022RfX,Nsch2019Colorful,Neto2021Multivariate}, boosting\,\cite{Liu2018Visual,Huang2019GBRTVis,Wang2021Investigating,Xia2021GBMVis}, or both ensemble learning methods~\cite{Chatzimparmpas2023VisRuler}. Most relevant to our work, VisRuler~\cite{Chatzimparmpas2023VisRuler} is a VA tool that \hl{assists users in making decisions based on Random Forest (RF) and AdaBoost models}. The tool's VA workflow involves selecting a diverse set of robust models, identifying important features, and determining essential decisions for global or local explanations. While our tool partially addresses the aforementioned challenges, our key focus is to obtain decision stumps with their assigned \emph{weights} and to enable \emph{rule overriding} of the resulting decision stumps extracted from an accurate and simpler AdaBoost model that approximates the behavior of a complex target model. These functionalities are both unsupported by VisRuler.

Several VA tools assist with the interpretation or diagnosis of the training process of gradient boosting models~\cite{Friedman2001Greedy}. GBMVis~\cite{Xia2021GBMVis} reveals the technical properties of gradient boosting, allowing the assessment of feature significance and decision-making tracking. BOOSTVis~\cite{Liu2018Visual} offers views such as a temporal confusion matrix, t-SNE projection~\cite{vanDerMaaten2008Visualizing}, and node-link diagram to monitor performance and examine rules. GBRTVis~\cite{Huang2019GBRTVis} uses continuous loss function monitoring to explore gradient boosting and visualizes the process with a node-link diagram and a treemap. \emph{VIS}TB~\cite{Wang2021Investigating} provides a redesigned temporal confusion matrix and feature impact comparison for per-instance prediction tracking, feature selection, and hyperparameter tuning. In contrast to \textsc{DeforestVis}, its node-link diagram designed for deep decision trees can limit the users' ability to evaluate many decisions simultaneously. Also, our choice of the AdaBoost algorithm (simpler than gradient boosting or other ensemble learning algorithms~\cite{Breiman2001Random,Wolpert1992Stacked,Chatzimparmpas2021StackGenVis}) to generate  decision trees in combination with a simple bar chart visualization allows users to instantly explore rules and compare the confidence of the surrogate model for each rule and feature.

Several VA tools have been developed to aid in the interpretation of RF models. iForest~\cite{Zhao2019iForest} shows the hierarchical structure of decision paths generated by RF. ExMatrix~\cite{Neto2021Explainable} uses a matrix-like visualization to analyze RF models and connect rules to classification results. Neto and Paulovich~\cite{Neto2021Multivariate} propose a tool for extracting and explaining patterns in high-dimensional data sets from random decision trees. Colorful trees~\cite{Nsch2019Colorful} uses a botanical metaphor to interactively explain the core parameters of RF models and allows for customized mappings of RF components to visual attributes. Finally, RfX~\cite{Eirich2022RfX} enables users to compare multiple decision trees from an RF model and manually adjust single  trees using overlapping histograms and dissimilarity projections. In contrast, \textsc{DeforestVis} helps the mining of rules with a focus on class outcomes for all and/or specific cases, provides a simple visual representation of the logic behind the produced rules, and retains the hierarchy of decision stumps due to the intrinsic AdaBoost's weighting system. In our work, users can explore the local and global impact of a manually overridden rule before confirming their action.

The visualization of single decision trees has been previously attempted through various methods such as node-link diagrams~\cite{Elzen2011BaobabView,Nguyen2000A,Lee2016An,Cavallo2019Clustrophile,Barlow2001Case,Phillips2017FFTrees,Bremm2011Interactive,Hongzhi2004Multiple,Munzner2003TreeJuxtaposer,Behrisch2014Feedback,Ware2001Interactive,Han2000RuleViz}, treemaps~\cite{Muhlbacher2018TreePOD,Gomez2013Visualizing}, icicle plots~\cite{Padua2014Interactive,Ankerst2000Towards}, star coordinates~\cite{Teoh2003Starclass,Teoh2003PaintingClass}, parallel coordinates~\cite{Tam2017An}, scatterplots~\cite{Marcilio2021ExplorerTree}, and scatterplot matrices~\cite{Do2007Towards}. However, these techniques do not work well when exploring multiple decision trees, which is important for understanding what individual trees have learned. Current visualizations of decision trees are not designed to explore model behavior. To address this, we propose a feature-aligned tree visualization that helps to understand and analyze rules across multiple one-level decision trees. Additionally, we summarize in segmented bar charts all decision stumps' predictive power and the final predictive outcome collectively.

Finally, to the best of our knowledge, no work in the literature describes the use of VA in conjunction with the AdaBoost ensemble learning algorithm~\cite{Schapire1999A} to generate interpretable decision stumps that aggregate the behavior of complex ML models.

\section{User Goals and Analytical Tasks} \label{sec:goals}
	We outline five \hl{User Goals (\textbf{UG1--UG5})} our and similar tools aim to achieve to extract easily comprehensible decision rules (\autoref{sec:userg}). Next, we identify five \hl{Analytical Tasks (\textbf{AT1--AT5})} that \textsc{DeforestVis} aims to help its users to complete (\autoref{sec:tasks}), following the guidelines from Munzner\,\cite{Munzner2009A}. These goals and tasks guide the design decisions made when developing \textsc{DeforestVis}. \hl{Throughout the process}, our target users are model developers and domain experts with a basic understanding of information visualization, a reasonable familiarity with the fundamental ML concepts, and a good understanding of their data. Our tool focuses on binary classification problems and tabular data with a limited number of meaningful features for the targeted users. We discuss these aspects further in Secs.~\ref{sec:intlim} and~\ref{sec:users}.

\subsection{User goals}
\label{sec:userg}
Our five user goals (described below) are more general and have been extracted from other relevant works which, overall, target similar users having
similar goals. Specifically, our five user goals were influenced by the research discussed in \autoref{sec:relwo},
the guidelines from Zhao et al.~\cite{Zhao2019iForest}, and our own experiences with interpretable/explainable ML~\cite{Chatzimparmpas2020A,Chatzimparmpas2020t,Chatzimparmpas2023VisRuler} and trustworthy ML~\cite{Chatzimparmpas2020The}. In particular, we took into account user goals and tasks outlined by Collaris and van Wijk~\cite{Collaris2022StrategyAtlas} in their study that involved interviewing six data science teams with an interest in explaining ML. Additionally, we considered the four user goals proposed by Antweiler and Fuchs~\cite{Antweiler2022Visualizing} based on their collaboration with healthcare professionals, which is also relevant for our usage scenario described in~\autoref{sec:case}.

\textbf{\hl{UG1: Replace an unintuitive ML model with an interpretable surrogate model for making decisions.}} As already outlined, our idea is to replace a complex and unintuitive model with an interpretable surrogate model that can approximate the original model's behavior while providing more transparent and understandable decision-making~\cite{Chatzimparmpas2020A,Chatzimparmpas2020The}. The surrogate model is a one-level surrogate decision tree or a rule-based system that will offer insights into how the model arrived at its predictions. When doing this, we also want that the used surrogate models should be easy to explore and understand by users (e.g., by avoiding deep decision trees---issues not addressed by prior works~\cite{Castro2019Surrogate,Ming2019RuleMatrix,Yuan2022Visual}).

\textbf{UG2: Identify good solutions for the trade-off between complexity and fidelity in approximation models.} To do this, it is necessary to carefully consider the specific problem at hand and the available resources, such as the time users are willing to spend and the free screen space~\cite{Chatzimparmpas2023VisRuler}. In some cases, a simple model with lower fidelity may be sufficient; in other cases, a more complex model with higher fidelity may be needed. Also, for surrogate decision trees, for each threshold value that decides if one instance falls into the left or right subtree, only a limited precision can be achieved~\cite{Chatzimparmpas2023VisRuler}. In contrast to previous works~\cite{Castro2019Surrogate,Ming2019RuleMatrix,Cao2020DRIL,Yuan2022Visual}, VA tools should communicate the impact of decimal precision to choose the appropriate rounding approach for the given problem. Finally, VA tools should support humans in reducing their cognitive load---as much as possible---while retaining high accuracy.

\textbf{UG3: Analysis of the extracted decision rules individually and jointly to understand the behavior of complex models.} Decision rules are a powerful tool to explain the behavior of complex ML models. By analyzing decision rules individually, we can identify which \emph{features are most important} in the model's decision-making process and how they are \emph{weighted}~\cite{Antweiler2022Visualizing}. Conversely, by examining decision rules jointly, we can gain a better understanding of how the model as a whole makes decisions and find any potential biases or limitations~\cite{Antweiler2022Visualizing}. Unlike RuleMatrix~\cite{Ming2019RuleMatrix}, the voting power of each weighted decision stump plays a vital role in surrogate models based on boosting algorithms, as is the AdaBoost method in our case. To sum up, VA tools enabling this exploration of standalone surrogate models can entirely replace a complex model if no further retraining processes are involved.

\textbf{UG4: Comparison of similarities/disparities over groups of training samples when changing a single decision rule.} The thresholds used for splitting in a decision tree can be adjusted based on prior knowledge of the problem domain~\cite{Antweiler2022Visualizing}. That is, a decision tree rule can be improved by repeatedly refining the split thresholds until they are properly adapted. However, optimizing decision trees in this way for each rule can result in models that are too specialized and may overfit the training data. Hence, it is important to test each rule with new training data to ensure that it is not overfitted. Similar to RfX~\cite{Eirich2022RfX}---but for surrogate models as is our case---we deem the adaptation of rules according to experts' prior knowledge an entry point for what-if hypotheses testing (see also \textbf{UG5}) and building their own trustful model~\cite{Yuan2022Visual}. VA tools should show how the adjustment of a particular threshold will influence the training instances that belong to different sub-branches of the decision stump.

\textbf{UG5: Multiple hypotheses about reversing the prediction for specific test cases.} Similar to iForest~\cite{Zhao2019iForest}---but for surrogate models as is our case---we want to assess the accuracy of a surrogate model's prediction \emph{vs} the original model for case-based reasoning~\cite{Collaris2022StrategyAtlas}. When the surrogate makes incorrect predictions for some test cases, it is crucial to find whether we can adjust a wrong prediction in the right direction. Formulating multiple hypotheses about reversing the prediction means considering several possible feature-based explanations of why the model wrongly predicted and assessing whether one can adjust it to make a correct prediction. VA tools should support this process by explaining to users what should be changed to get a correct prediction or, more generally, adjust the surrogate model according to the domain expert's prior knowledge (as described in \textbf{UG4}).

\begin{figure*}[tb]
\centering
\includegraphics[width=\linewidth]{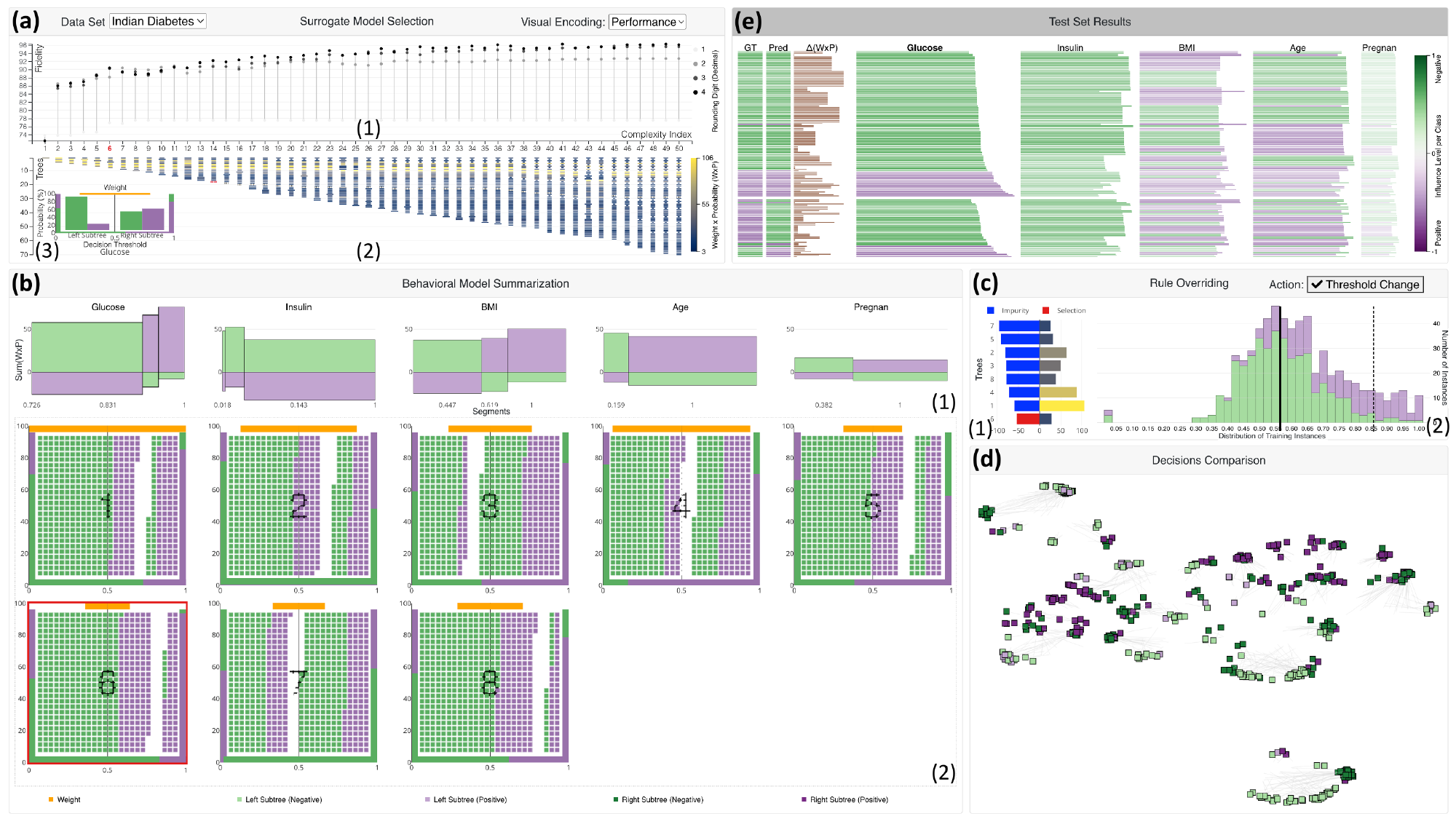}\vspace{-2mm}
\caption{\hl{Components of \textsc{DeforestVis}: (a.1) lollipop plot shows data-rounding effects in the fidelity score for four different decimal digit precisions; (a.2) dot plot with lines of various widths (unique rules/stumps $>$ original $>$ duplicated) and colors (visual encoding: performance, that is, weighted probability ($W \times P$)) that explains complexity increase as more decision stumps get added and (a.3) selective stump-based explanation; (b.1) segmented bar chart tells the predictive outcome and power of each segment based on automatically computed thresholds and (b.2) detailed stump-based explanation grid; (c.1) bar chart shows the impurity and weighted probability of each decision stump; (c.2) histogram shows the active rule's threshold and distribution of training instances; (d) projection aggregates the global behavior of instances; color shows the local behavior according to the currently selected decision stump; and (e) fragmented bar chart shows the per-feature contribution and influence level for each test case. The main questions that the five views of \textsc{DeforestVis} address are: (a) Which surrogate model gives the user the desired fidelity/complexity trade-off (\autoref{sec:surrrogate})? (b) How do feature thresholds affect the selected surrogate model, summarizing the behavior of the complex model (\autoref{sec:behavior})? (c) How can specific rules be overridden (\autoref{sec:rule})? (d) What changes due to such user actions (\autoref{sec:decisions})? (e) How does the new surrogate model perform on unseen test data (\autoref{sec:test})?}}
\label{fig:teaser}
\end{figure*}

\subsection{Analytical tasks} \label{sec:tasks}
Considering the guidelines from Munzner\,\cite{Munzner2009A}, we found five analytical tasks concerning concrete operations that users should be able to perform using
\textsc{DeforestVis} to achieve our user goals.

\textbf{AT1: Use shallow decision trees to split the complex model and see the impact of information reduction in faithfulness.} Users should be able to see how the precision value picked for the threshold in each shallow decision tree affects the accuracy of the whole surrogate model (\textbf{UG1}). Rounding up a few threshold-decimal values for each decision stump will reduce the time required for users to grasp the value of deciding in favor of one or the other class.

\textbf{AT2: Find the \cit{optimal} number of decision trees needed to retain high-enough model performance.} Following \textbf{AT1}, users should be guided through the process of selecting the appropriate number of decision trees that preserves high enough fidelity for their given problem (\textbf{UG2}). One drawback of global surrogate models is that they cannot confidently tell how close to the target model is `enough' for a selected surrogate to be. This is a decision that ultimately is to be taken by users that explore such surrogates for a given concrete problem\,\cite{Molnar2020Interpretable,Castro2019Surrogate}. Thus, it is crucial to enable the comparison of the predictive accuracy of surrogate models having different numbers of decision trees. Few trees are easier to understand but likely less precise. Many trees are arguably problematic since users are unlikely to engage meaningfully with hundreds of rules.

\textbf{AT3: Examine the summarized thresholds for each feature that lead to different predictions and drill down to investigate single decision rules.} The summarization of the per-feature decisions in a single view that combines the decisions sorted from the most to the least contributing features allows users to assess the influence of each feature (\textbf{UG3}). Understanding the thresholds and decision rules can help users to explain the model's predictions. Once we have a clear understanding of these thresholds, we can drill down to investigate the underlying single decision trees/rules that the model combines to arrive at its final prediction. This involves enabling users to examine each individual feature and the threshold value associated with it.

\textbf{AT4: Assist users in manually adjusting decision rules and provide visual feedback about the impact of their actions.} Manually adjusting rules is useful when the default rules do not fit users' needs or when the system's performance needs to be optimized based on expert decisions (cf. \textbf{UG4}). By allowing users to adjust each decision tree/rule, they can customize the application to their specific use case or preferences. The impact of such changes locally for a specific rule should be juxtaposed to the global influence on all decision rules and for an entire training and/or test set.

\textbf{AT5: Experiment with what-if scenarios to predict particular test instances in a different class.} An aggregated explanation of why specific test cases were misclassified should be highlighted for users (\textbf{UG5}). They should be capable to explore borderline cases that are close to the decision boundary of a class and form new hypotheses about which threshold one should change to classify such cases into another class. The influence level of each feature and the ease of manipulating a feature should both be visualized for users to evaluate.

\section{DeforestVis: System Overview} \label{sec:overview}
	\begin{figure}[t]
\centering
\includegraphics[width=\linewidth]{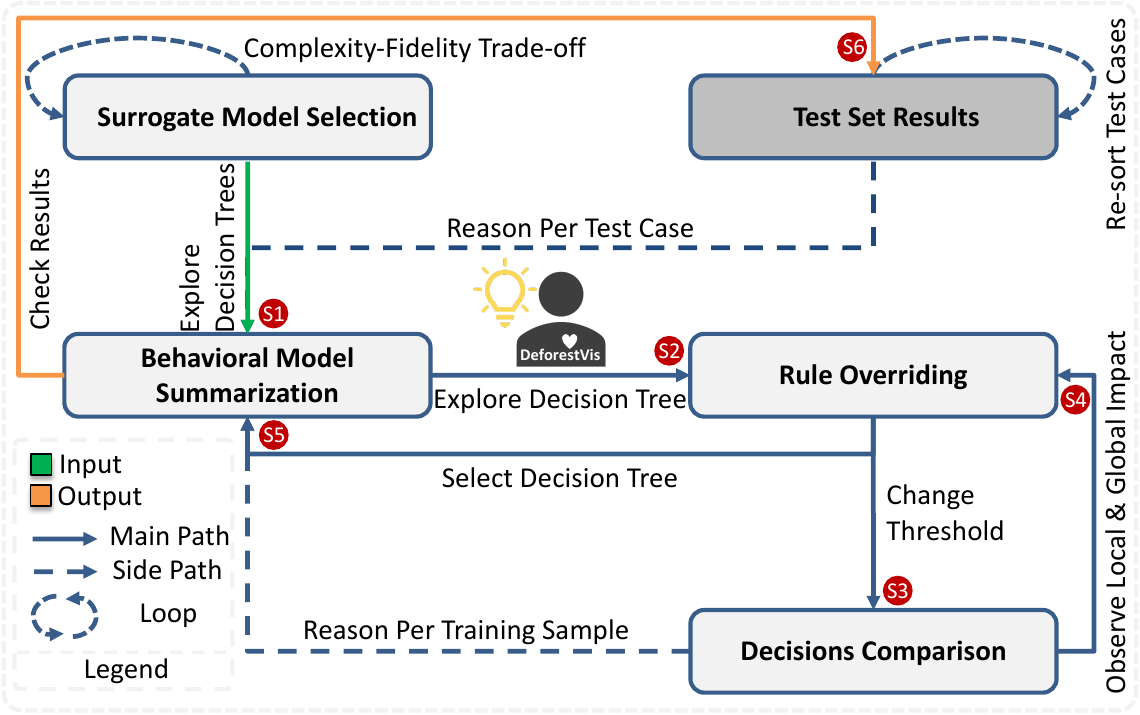}\vspace{-2mm}
\caption{The \textsc{DeforestVis} workflow enables users to choose the preferable surrogate model according to their willingness to sacrifice fidelity in favor of less complexity, explore the decisions extracted from the surrogate model (which serves as a simplified representation of the complex ML model), and manipulate individual rules based on the visual feedback and their prior experiences. To close the loop, users can reason about specific test cases iteratively while exploring the already-existing exported explanations for the training data. Steps \circled{S1}--\circled{S6} showcase a simple single-iteration loop.}
\label{fig:workflow-diagram} %\vspace{-1.5mm}
\end{figure}

\begin{figure*}[tb]
\centering
\includegraphics[width=\linewidth]{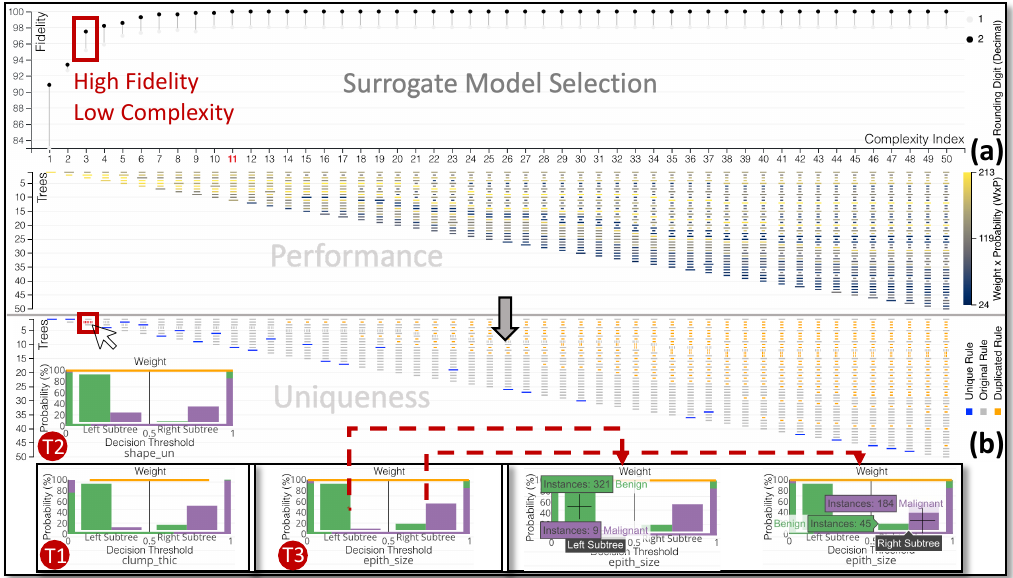}\vspace{-2mm}
\caption{Exploration of the complexity-fidelity trade-off with \textsc{DeforestVis}. The lollipop plot (a) shows the \emph{fidelity score} against the \emph{complexity index} for the incremental increase in \emph{number of estimators} hyperparameter of the surrogate models created by AdaBoost. Rounding decimal digits that could result in information loss are also visible here. The highlighted surrogate model has only three one-level decision trees but can emulate over 97\% the target model. The dot plot with lines in the same view shows the contribution of each decision to the final result. The higher the value of the \emph{weighted probability (weight $W$ $\times$ probability $P$)}, the more important a rule is. View (b) shows an alternative encoding with newly discovered decision stumps having larger  width and colored blue. Already found rules which are still part of the next model are in gray; after they are included multiple times, they become duplicated rules (orange, smallest width). The user clicks a rule of interest to inspect (red).}
\label{fig:use_case1_surrogate}
\end{figure*}

% \begin{figure*}[t]
%   \centering
%   \includegraphics[width=\linewidth]{usecase1_models_S2-crop}\vspace{-2.5mm}
%   \caption{Examining the behavioral summary of the best surrogate model. In (a), the user explores the most important feature that suggests benign cancer for less than 0.35 and malignant cancer for the rest with diverse probabilities. The user hovers over a segment of conflict in \circled{S1} that triggers the system to highlight the relevant parts for the two decision stumps composing the overarching decision. The modification of each rule is supported in (b), where the default rule has the highest impurity and the lowest weight (just in this case). \circled{S2} denotes the users' selection of the more powerful rule from the two related to \emph{bare\_nuc} feature. The distribution of training data in the histogram is visible along with the threshold after the choice of rule \#1.}
%   \label{fig:use_case1_summarization}
% \end{figure*}

\begin{figure*}[tb]
  \centering
  \includegraphics[width=1.0\linewidth]{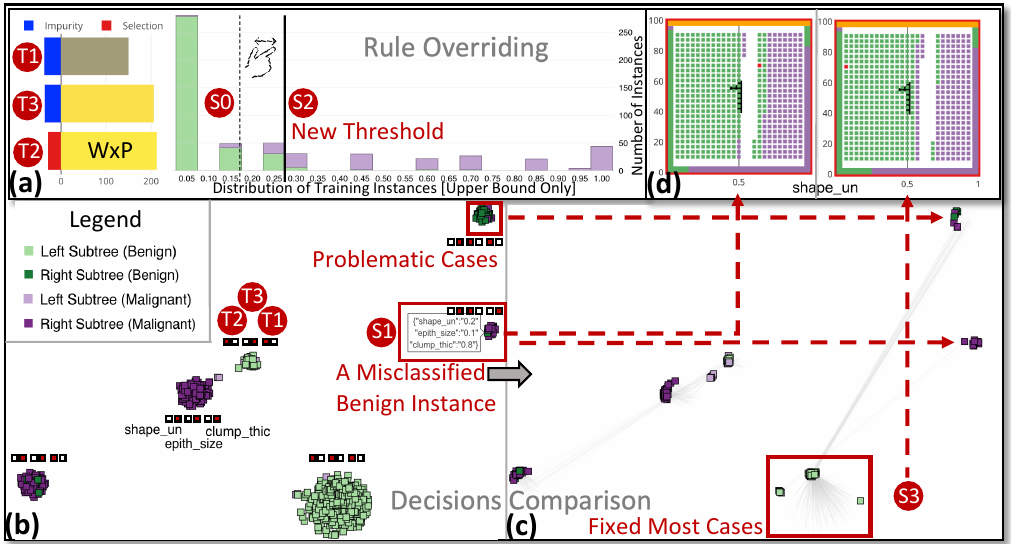}\vspace{-2mm}
  \caption{Impact of manually overriding the most-contributing rule in a low-complexity (index \#3)  but high enough fidelity scenario (shown  in~\autoref{fig:use_case1_surrogate}(b)). In (a), the user clicks on rule \circled{T2} (cf. \autoref{fig:use_case1_surrogate}) and, after studying the distribution of training samples on each side of the subtree,  decides to adjust the threshold (from \circled{S0} to \circled{S2}). The initial state (b) has some problematic cases, marked \circled{S1}, belonging to the right subtree of \circled{T2}, left subtree of \circled{T3}, and right subtree of \circled{T1} based on the manual inspection by hovering over each of them (see small red boxes). These are benign samples mixed with malignant ones. Even though the threshold change \circled{S2} adversely affects the classification of malignant cases, its benefit overcomes the default suggestion. View (c) confirms this: the trajectory of points shows a better separation of the previously highlighted sample groups. After the change \circled{S2}, the hovered point (among others) has moved from the right subtree to the left, which is correct (see (d)).}
  \label{fig:use_case1_rule}
  
\end{figure*}

\begin{figure*}[t]
  \centering
  \includegraphics[width=1.0\linewidth]{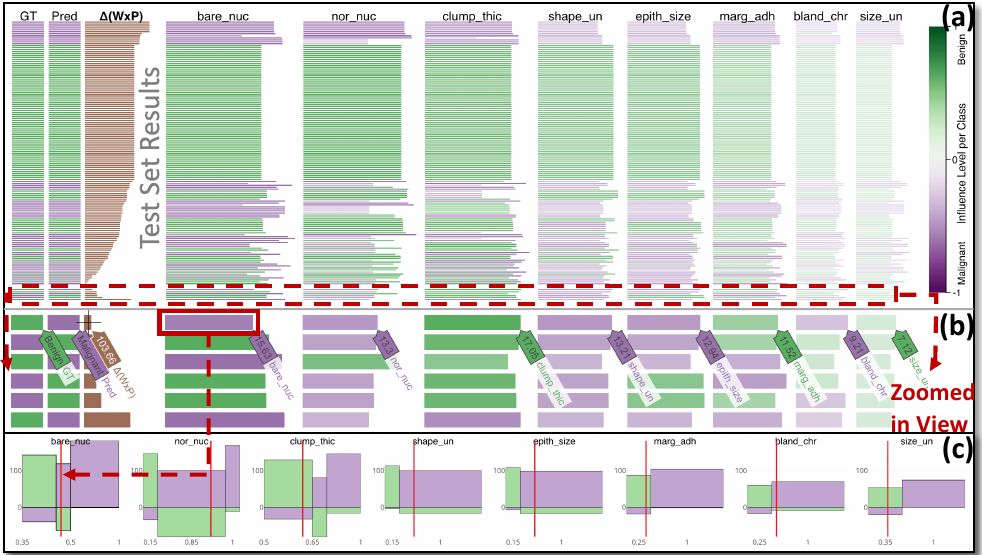}\vspace{-2mm}
  \caption{Local analysis uses the most accurate (and as simple as possible) surrogate model (complexity index \#11, \autoref{fig:use_case1_surrogate}(a)) to predict a test case. (a) shows the certainty of the surrogate model classifying the test set. Test cases at the bottom are misclassified, with the easiest-to-swap class instance drawing the user's focus, see (b). We want to find the feature that contributes the most to this case being misclassified as malignant. After \emph{clump\_thic}, which moves the final prediction toward the benign class, the next most contributing feature is \emph{bare\_nuc} based on the length of the bars and the strong opaque color revealing the influence level of the features. A threshold increase from $\approx$0.35 to $\approx$0.45 would swap the prediction for this test case, as shown in (c). Yet, this fluctuation can harm generalizability since some malignant cases would fall in the left/wrong subtree.}
  \label{fig:use_case1_results}
  %\vspace{-1em}
\end{figure*}

\begin{figure*}[tb]
  \centering
  \includegraphics[width=1.0\linewidth]{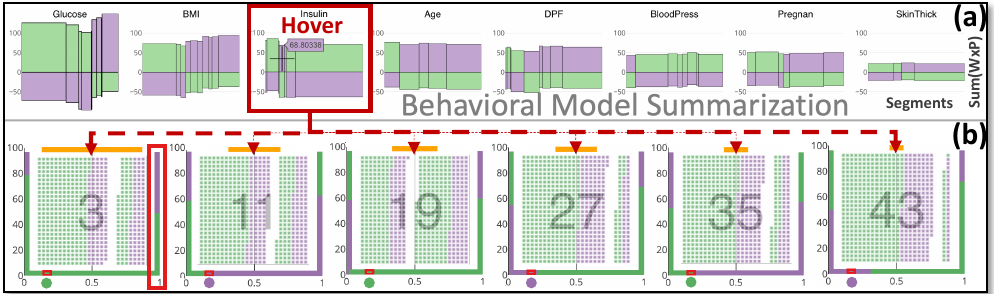}\vspace{-2mm}
  \caption{Analysis of decision stumps explaining how slight \emph{insulin} amounts classify patients as diabetic. Hovering over this feature (a) allows Amy to follow the path of how this aggregated rule was created. To Amy's surprise, the model appears confused in (b) as there is a balance between the six present decision stumps---three suggest the positive class,  the rest suggest the opposite. The largest-weight rule seems divided since the probability of its right subtree is 50\% (red box), so it eliminates itself from the prediction outcome. The remaining rules make the decision. The second more impactful rule favors the negative class. Given this analysis, Amy could modify the rules to change this strange behavior (see~\autoref{sec:case}).}
  \label{fig:use_case2_summarization}
  \vspace{0.5em}
\end{figure*}

 We have developed \textsc{DeforestVis}, an interactive web-based VA tool that allows users to explore the behavior of complex ML models with feature-based explanations from surrogate decision stumps to meet our user goals and analysis tasks (\autoref{sec:goals}). The frontend of \textsc{DeforestVis} is developed in JavaScript using Vue.js~\cite{vuejs}, D3.js~\cite{D3}, and Plotly.js~\cite{plotly}; the backend is written in Python using Flask~\cite{Flask} and Scikit-Learn~\cite{Pedregosa2011Scikit}.

\noindent\textbf{Views and workflow:} To keep---as small as possible---the number of views needed for a system that visualizes surrogate models, \textsc{DeforestVis} has five main views (\autoref{fig:teaser}): (a) surrogate model selection ($\rightarrow$ \textbf{AT1} and \textbf{AT2}), (b) behavioral model summarization ($\rightarrow$ \textbf{AT3}), (c) rule overriding, (d) comparing decisions ($\rightarrow$ \textbf{AT4}), and (e) test set results ($\rightarrow$ \textbf{AT5}). These views support our workflow in \autoref{fig:workflow-diagram}: (i) build several surrogate models with increasing complexity by including more decision stumps in architectures (\autoref{fig:teaser}(a));
(ii) select a surrogate model with low-complexity and high-fidelity to fit the one's desired precision (\autoref{fig:use_case1_surrogate}(a));
(iii) analyze the behavior of the target model by exploring the summarized \emph{decision threshold} per feature from the weighted decision stumps (\autoref{fig:teaser}(b));
(iv) examine both local and global impact of overriding an automatically produced rule by comparing decisions while adjusting the threshold value (\autoref{fig:teaser}(c) and (d)); and
(v) observe the influence of the final surrogate model on unseen data and optionally reason why a test case was classified as a given class based on the contribution of each feature (\autoref{fig:teaser}(e)).
By repeating the workflow in~\autoref{fig:workflow-diagram}, the user gains knowledge about what the target model has learned from the data and, in addition, can fine-tune the target model's decision-making via the surrogate model.

\noindent\textbf{Implementation details:} \textsc{DeforestVis} uses the state-of-the-art ensemble learning Explainable Boosting Machine (EBM) approach~\cite{Nori2019InterpretML}. Yet, our workflow is model-agnostic since AdaBoost-based surrogate models can approximate the behavior of any ML model. We chose EBM intentionally because this algorithm produces systematically fewer decision rules compared to other ensemble learning methods that we have experimented with, e.g.,  XGBoost~\cite{Chen2016XGBoost} or Random Forest~\cite{Breiman2001Random}. For all our experiments (including the use cases in~\autoref{sec:case}), we use EBM with the default hyperparameters as the target model, and we further split data into 80\% training and 20\% testing with a stratified strategy (i.e., keeping the same balance in all classes for both sets). We randomly sample the hyperparameter space (50 Random Search iterations) in order to visualize each AdaBoost model with an increasingly larger number of decision stumps (\emph{n\_estimators} hyperparameter). We used the default AdaBoost hyperparameters~\cite{Schapire1999A} except for the maximum number of features (\emph{max\_features}) to use when looking for the best split, which we set to the square root of the number of features. For more technical implementation details, we refer the reader to our source code repository~\cite{DeforestVisCode}. We describe \textsc{DeforestVis} by an example with the \emph{breast cancer (Wisconsin)} data set\,\cite{Dua2017} (699 samples, 9 features, 2-class classification task:\,\emph{benign}, \emph{malignant}). All our data sets are normalized to $[0,1]$.

\subsection{Surrogate model selection} \label{sec:surrrogate}
After creating 50 surrogate models by gradually increasing complexity, we use \textsc{DeforestVis} to show their training-prediction accuracy with a lollipop plot ($y$-axis: fidelity; $x$-axis: complexity; see \autoref{fig:teaser}(a.1); \textbf{AT1}). In \textsc{DeforestVis}, \emph{fidelity} is defined as the \emph{accuracy} with which every surrogate model can simulate the target model~\cite{Castro2019Surrogate}, while \emph{complexity} is the \emph{number of decision stumps} in each surrogate model (\textbf{AT2}). The top circles in the plot show threshold precision---defined as how much rounding decimal digits affects the surrogate's fidelity---via a grayscale colormap (light gray: less precision; black: maximum precision; \textbf{AT1}).

The dot plot with lines in \autoref{fig:teaser}(a.2) below the lollipop plot shares the same $x$-axis as the lollipop plot. Its lines show decision stumps included in every surrogate model (\textbf{AT2}). These are colored to show either (1) performance (using the colorblind-friendly version of Viridis colormap, see \autoref{fig:teaser}(a.2)) or (2) the uniqueness of a decision stump (see \autoref{fig:use_case1_surrogate}(b)). The first option encodes the weight $W$ of each rule multiplied by the predicted probability $P$ of all instances to belong to the Ground Truth (GT) class. The second option scans the space of surrogate models from fewer to more decision stumps being produced. Longest, blue lines are unique decision stumps (or rules); \cit{original} rules found in an earlier smaller surrogate model which exist just once in the current surrogate model are gray; and duplicated rules found twice or more in the same surrogate model are orange, narrow lines.
Next, users can choose between the two visual encoding modes and click on a line to bring up a pop-up with the decision stump/rule it encodes on the left-hand side of the dot plot (\autoref{fig:teaser}(a.3) and \autoref{fig:use_case1_surrogate}(b))). \hl{When doing this, the solid lines encoding the same (in terms of feature and threshold) decision stump in all surrogate models change to dashed lines, with the most-common decision stumps globally located at the top of each stack of decision stumps.} This lets users quickly explore \emph{unique and influential} decision stumps that \emph{should} be part of their surrogate model, and choose the appropriate stumps for their surrogate model. Another use case is to get inspiration from a more complex surrogate model on how to modify a decision stump of a less complex surrogate model. Users can next select another surrogate model than the default (highest-fidelity, lowest-complexity) one to analyze. The currently selected surrogate model (\autoref{fig:use_case1_surrogate}(a), complexity index \#11) gets its index label marked red.

We compactly visualize a decision stump (\autoref{fig:use_case1_surrogate}(b), \circled{T2}) by a four-component visual design that shows all relevant information of AdaBoost's one-level decision trees. Each stump uses a single feature to cut the training instances into two subtrees. The top orange bar expands left and right from the middle to show the weight (influence) of a single AdaBoost stump. In the remaining visual components, green maps one class and purple the other (in our case benign and malignant cancer, respectively). The bottom bar shows the \emph{decision threshold value} (range 0 to 1) that separates training instances below that threshold to either class; in our case, this threshold is around 0.15---below this value, \circled{T2} predicts the green class. The left stacked bar on the boundary shows the predicted probability (or confidence) with which the decision tree suggests that the training samples of the left subtree are in one or the other class. For \circled{T2}, almost 100\% probability is in favor of the green class; for the right subtree, approx.~80\% suggests the purple class and the remainder the green class. The bar chart in the middle shows the distribution of training samples (i.e., the exact number of instances) that fall into the left or right subtree (color encodes GT class, see above). In \circled{T2}, most training samples of the left and right subtrees are correctly predicted as the green, respectively purple, classes.

\noindent\hl{\textbf{Design discussion.} An alternative to the lollipop and dot plots shown in~\autoref{fig:teaser}(a) is using a scatterplot to visualize the balance between fidelity and complexity, as suggested in~\cite{Muhlbacher2018TreePOD}. This method could also help users to seek solutions on the Pareto frontier. A scatterplot might also scale better with the sample count. However, a scatterplot has a few drawbacks: (1) it can complicate the selection and exploration of \emph{individual} decision stumps; (2) it could obscure the sequential relationship between stumps; and (3) it may not effectively show the rounding effect on the surrogate model's accuracy and require additional designs to show such information.}

\noindent\hl{\textbf{Design scalability.} To improve the scalability of \textsc{DeforestVis}, a solution is to separate the visualizations in~\autoref{fig:teaser}(a) into a different tab specifically conceived for selecting the optimal surrogate model. Additionally, our choice of representing a decision stump by a four-component visual design is to save space and encode diverse information compared to, for example, more traditional node-link diagrams.}

\subsection{Behavioral model summarization} \label{sec:behavior}
This view includes a grid of stumps at the bottom that is summarized with a segmented bar chart on top, and it is the most important view that intentionally has the largest size because the other views are \emph{connected} with this behavioral model summarization central view.

The decision stumps in~\autoref{fig:teaser}(b.2) follow the same design as the active stumps explained before, except we replace the bar chart with a color-coded grid showing training samples. The goal behind the active stumps is to get inspired from other more complex surrogate models to override some of the decision stumps of the selected surrogate model accordingly. In contrast, the grid we use here enables the in-depth exploration of which subtree each sample falls into (\textbf{AT3}). Every grid contains two cell groups, one to the left, one to the right, ordered top-to-bottom by the predicted probability of each sample belonging to the GT class, both groups separated by some whitespace in the middle of the grid. Misclassified samples are shown as cells closer to the grid middle, with colors mapping the GT of each training sample. Each feature is encoded by a series of decision stumps, with the most important features (having a higher sum of weighted probability for all decision stumps jointly influencing them) 
listed first, e.g., \emph{Glucose} and \emph{Insulin} in~\autoref{fig:teaser}(b.1). 

For each feature, the top part of \autoref{fig:teaser}(b.1) shows a segmented bar chart summarizing all its decision stumps, with one segment (rectangle) added per threshold value (\textbf{AT3}). For example, if two decision stumps exist with different thresholds, the segmented bar chart will be split into three rectangular segments, while for one stump, the information represented is equivalent to the decision stump itself. Therefore, the chart's $x$-axis encodes the all available thresholds that separate predictions into different classes; the $y$-axis shows weighted probability for the most probable class above the zero line, and weighted probability for the other class on the negative side (i.e., rectangles below zero). When hovering over a segment, the corresponding decision stump is highlighted to show how, and with what magnitude, the stump votes. The most impure decision stump (based on the \emph{Gini impurity measurement}~\cite{Genuer2010Variable}) is selected by default and marked in red. High impurity is problematic because it indicates a poor separation between classes.

\noindent\hl{\textbf{Design discussion.} Although the detailed stump-based explanation grid may initially seem complex, it is essentially a compact visualization for each decision stump, depicting the (1) threshold for each decision, (2) predicted probability for the left condition, (3) predicted probability for the right condition, (4) weight of each stump, and (5) training instances with the GT class. Despite its complexity, the grid can be easily understood by breaking it down into five \emph{design zones}. Also, the visual encoding and color scales shared by this grid and the projection view help users decode the grid visual representation.}

\noindent\hl{\textbf{Design scalability.} While our current grid design is quite functional, its scalability to large data sets might be challenging. To overcome scalability issues regarding the display of many instances in the decision stumps (\autoref{fig:teaser}(b.2)), \textsc{DeforestVis} reduces each box to plot each training sample by a single pixel. A benefit of this chosen fine-grained grid design (of raw data) is the sorting of instances
according to how easy or hard it is for an instance to be classified in the opposite class. However, if more samples exist than available pixels, we could aggregate or group similar samples into larger boxes\,\cite{Elmqvist2010Hierarchical,Munzner2014visualization}. While fully scalable to any instance count, this solution would make it harder to drill down to single problematic training instances. In the extreme case, when having thousands of instances and features, this view could be completely replaced by the segmented bar chart. We suggest further solutions to this problem and other potential limitations found by the experts in~\autoref{sec:eval}.}

\subsection{Rule overriding} \label{sec:rule}
The bar chart (\autoref{fig:teaser}(c.1)) shows impurities with the weighted probability score for each stump on a colorblind-friendly Viridis colormap once more (\autoref{sec:surrrogate}; \textbf{AT4}). The $y$-axis shows the identification index of the stumps with rules, highest impurity at the top. The currently selected stump is marked red (\autoref{sec:behavior}). In \autoref{fig:teaser}(c.1), for instance, the \nth{6} stump has the lowest impurity value (close to -50) but its weighted probability score is low, see its dark blue color and short bar length.

The histogram in~\autoref{fig:teaser}(c.2) groups training instances into 10 or 20 bins, depending on the user-chosen decimal precision. When more precision is desired, e.g. \autoref{fig:teaser}(c.2), the highest value among the bins is used (\textbf{AT4}). The dashed black vertical lines show the threshold value before, respectively after, the user's interaction with the currently active decision stump (marked red). GT classes are color-coded in green and purple, like in the other views.

\subsection{Comparing decisions} \label{sec:decisions}
\autoref{fig:teaser}(d) helps to study further one's change of a rule's threshold value. We use here a UMAP projection\,\cite{McInnes2018UMAP} of all training samples (\textbf{AT4}). Samples are colored by class with brightness mapping the local relationship of these samples with the \nth{6} decision stump that is the purest one and relates to the most important feature, i.e., \emph{Glucose}. Bright and dark colors show samples in the left, respectively right, subtree of the selected decision stump only (i.e., local investigation). Finally, each input dimension taken for UMAP encodes one decision stump of the investigated surrogate model and equals 0 for samples in the left subtree of the stump, else 1. 
In~\autoref{fig:teaser}(b.2), we thus have 8 dimensions since the studied surrogate model has 8 stumps that are being reduced to 2 dimensions (i.e., global investigation). As such, samples identically classified by all stumps will be positioned close to each other.

When users change the threshold value, the projection updates. We show how points move in the projection (before \emph{vs} after the change) by lines, so users can understand how their changes in one rule impact specific samples \emph{locally} for a specific stump and \emph{globally} for all sample pairs falling into different subtrees of all stumps (\textbf{AT4}).

\noindent\hl{\textbf{Design discussion.} For the UMAP projection, we have tried using shape instead of color for visualizing if a sample belongs to the left or right subtree of the selected decision stump in earlier design iterations of \textsc{DeforestVis}. But, with
that design, it was almost impossible to correctly see if instances belong to the left or right subtree.}

\noindent\hl{\textbf{Design scalability.} To limit overplotting in a projection, Collaris and van Wijk~\cite{Collaris2022StrategyAtlas} recommended a restriction to a maximum of 5,000 instances. This indication can also be applied to our UMAP projection view. We relax this limit by allowing zooming and panning our projection view. The use of techniques to reduce clutter in projections and edge bundling for the lines showing the trajectory of points can further improve overplotting~\cite{Hilasaca2019Overlap,Yuan2021Evaluation}.}

\subsection{Test set results} \label{sec:test}
To test if the user's changes do not overfit the surrogate model, our VA tool shows the ground truth (\emph{GT}) and predicted (\emph{Pred}) results for every test sample (\autoref{fig:teaser}(e) and~\autoref{fig:use_case1_results}(a); \textbf{AT5}), with each table row being a test case. The brown $\Delta(W \times P)$ column shows the difference in weighted probability needed to switch the prediction from one class to the other. A low $\Delta(W \times P)$ value means low classification confidence for that particular test sample. The bar charts to the right of $\Delta(W \times P)$ show the contributions of each feature proportional to all features (summing up to $100\%$ pixel length of the view; see the values in~\autoref{fig:use_case1_results}(b)) for the prediction of each test sample. Color encodes the predicted class (green or purple). \hl{Color saturation double-encodes the \emph{influence level per class} to highlight the most impactful features for a given test instance, with fully opaque bars displaying high confidence, suggesting that the weighted probability is higher for those features than others (\textbf{AT5}).}

\noindent\hl{\textbf{Design discussion.} If perceptual complexity was not an issue for the bar charts to the right of $\Delta(W \times P)$, we could alternatively use color saturation to map the prediction boundary difference. For example, a fully opaque bar would mean small adjustments to the decision threshold of the decision stump(s) relevant for this feature will lead to a change in the predicted class outcome (serving a similar purpose as the red lines in~\autoref{fig:use_case1_results}(c)). In summary, our VA tool leverages simple yet effective visualizations that ML experts are already accustomed to, such as bar charts, histograms, and projections. This was also confirmed by the ML experts from our interview sessions, as demonstrated in~\autoref{sec:eval}. By using such familiar techniques, we aim to minimize the learning curve for new users.}

\noindent\hl{\textbf{Design scalability.} If the table's scalability becomes an issue, a solution is to group similar samples and visualize only the misclassified test instances~\cite{Elmqvist2010Hierarchical,Munzner2014visualization}. However, the current granularity level of \textsc{DeforestVis} has many benefits, including being
able to explain \emph{specific test instances} and trace back \emph{raw data instances} to the surrogate model’s internal parts.}

%The columns to the right of $\Delta(W \times P)$ show the contributions of each feature to the prediction of each sample as bar charts---color encodes the predicted class (green or purple). Color saturation shows the prediction confidence---fully opaque bars show high confidence, meaning that large adjustments of the thresholds of the decision stumps relevant to that feature are needed to change the predicted class (T5).

\section{Use Cases} \label{sec:case}
	We next present a use case and a usage scenario showing how \textsc{DeforestVis} evaluates the behavior and summarizes the knowledge generated from a complex ML model.

\subsection{Use case: Additional analysis support} \label{sec:application}

We next discuss additional exploration features of our VA tool, using the same data set as in~\autoref{sec:overview}. Although tools like RuleMatrix~\cite{Ming2019RuleMatrix}, DRIL~\cite{Cao2020DRIL}, and the one designed by Di Castro and Bertini~\cite{Castro2019Surrogate} (see~\autoref{sec:relwo}) provide feature importance and tree exploration insights similar to \textsc{DeforestVis}, our proposed tool allows users to vary the complexity level of the surrogate model in different ways to explain with only a few trees an entire complex model, resulting in an adaptive design that may be more practical to real-world scenarios, as motivated in~\autoref{sec:intro}, supported by Yuan et al.~\cite{Yuan2022Visual}, and the ML experts in~\autoref{sec:eval}.

\textbf{Exploring dynamics of different surrogate models.} In the scenario presented so far to illustrate the tool's views, we observe from~\autoref{fig:use_case1_surrogate}(a) that rounding the thresholds of the decision stumps to two decimals leads to the highest possible fidelity in all 50 surrogate models, which already to a certain extent minimizes the users' cognitive load (\textbf{AT1}). The surrogate model with complexity index \#11 (\autoref{fig:use_case1_surrogate}(a), marked red), containing 11 decision stumps, already explains 100\% of the original model's behavior. In the dot plot with lines of~\autoref{fig:use_case1_surrogate}(a), we see that the weighted probability value for many newly produced/unique rules (for surrogate models after model \#11) substantially decreases to $\approx$24 (indicated by dark blue colors appearing for the unique rules with lengthy bars in~\autoref{fig:use_case1_surrogate}(a)). We next choose to explore a simplified surrogate model (the selection is marked with the blue box in~\autoref{fig:use_case1_surrogate}(b)) composed of only three decision stumps since this reduces complexity drastically (by almost 73\% if we count the number of decision stumps in each surrogate model) and fidelity remains above 97\% (\textbf{AT2}). We select the unique decision stump (identifiable by its larger width) introduced in this specific surrogate model---marked with dashed red lines in~\autoref{fig:use_case1_surrogate}(b) (\textbf{AT3}). This stump has the maximum possible weight (see the lengthy orange bar on top of \circled{T2}), thus it is a \emph{core rule} for the \emph{shape\_un} feature. The pop-up \circled{T2} tells that, when the value is lower than 0.15, the probability of samples being classified as benign is very high according to the left subtree. \circled{T1} has less impact due to its smaller weight, but it classifies malignant (purple) samples with very high probability when \emph{clump\_thic} is above $\approx$0.6 (indicated by the purple color at the bottom bar of the \circled{T1}). Another interesting decision stump is \circled{T3}, with 321 training samples belonging to the benign class and 9 misclassified in this subtree with extremely high predicted probability and confidence (see the tree weight). The right subtree of \circled{T3} has 184 malignant instances and 45 benign instances with a probability above 80\% (compare the purple bar to the green bar on the right-hand side).

\textbf{Analyzing and overriding a decision rule.} We move on to a deeper exploration of the surrogate model with three decision stumps. In~\autoref{fig:use_case1_rule}(a), we selected the purest, most impactful, decision stump (\textbf{AT4}). In step \circled{S0}, we see that the automatically-generated stump finds values below 0.15 as benign while the GT in the histogram tells us that the bin between 0.2 and 0.25 has more benign than malignant cases. To confirm this, we study the projection of training samples forming different groups (\autoref{fig:use_case1_rule}(b)), looking whether the samples belong to the three decision stumps (\circled{T2}, \circled{T3}, and \circled{T1}). The problematic (confused) cases are marked by red boxes, with many benign cases being left out in the right subtree instead of the left. One such benign case is marked in \circled{S1} with \emph{shape\_un} being 0.2. To handle such cases, we change the threshold to a new value of 0.25, step \circled{S2}. This makes the benign (dark green) cases light green and move to the benign cluster (\circled{S3}, \autoref{fig:use_case1_rule}(c) bottom). The highlighted case visible in~\autoref{fig:use_case1_rule}(d) has been resolved by this action and moved from the right subtree to the left one (\textbf{AT4}). 

\textbf{Testing a borderline test case hypothesis.} In~\autoref{fig:use_case1_results}(a), we spot test cases that are misclassified by the surrogate model (\textbf{AT5}). How easy it is to change the prediction for borderline cases with the lowest $\Delta(W \times P)$ visible in~\autoref{fig:use_case1_results}(b)? The highest-contribution feature is \emph{clump\_thic} (longest bar in~\autoref{fig:use_case1_results}(b) for the hovered test case) which influences the result in the correct direction (green color). The second most-contributing feature is \emph{bare\_nuc}, 15.63\% of the sum of weighted probability for all features. By hovering over this test case, we see that its \emph{bare\_nuc} value is marginally in favor of the purple (malignant) class with a 0.35 threshold (\autoref{fig:use_case1_results}(c)). Adjusting the rule to 0.45 would lead to this test case falling into the green class but could negatively affect other training or test samples. To solve this, we can repeat the analyzing and overriding of a decision rule procedure described earlier above.

\subsection{Usage scenario: Behavioral summary of the target model} \label{sec:scenario}

Amy---a data analyst in a hospital---got a labeled data set on  \emph{diabetes}\,\cite{Smith1988Using} with 8 features, 768 samples. She splits the data set into 80\% training and 20\% test samples. Amy aims to fit a highly accurate (but complex) ML model to the training data and check its prediction ability on the test set. Yet, from her experience, she knows that it is hard to check what complex ML models do learn. She uses \textsc{DeforestVis} to analyze if such a model performs well and also presents to the doctors the main findings using a simpler surrogate model that facilitates their domain expertise injection on the decision rules. Her end goal is to \emph{fully replace} the target model.

\textbf{Inspecting an unusual feature of interest.}
Amy begins her exploration with the default surrogate model created by \textsc{DeforestVis}, which achieves the highest possible fidelity of approx. 96\% (\textbf{AT1}). The surrogate model index \#41 contains many decision stumps (\autoref{fig:teaser}(a.1)), so interpreting the model's behavior is hard (\textbf{AT2}). However, \textsc{DeforestVis} provides a summary of the predictions and confidence levels for all decision stumps extracted for each feature, see~\autoref{fig:use_case2_summarization}(a) (\textbf{AT3}). Amy quickly notices that \emph{Glucose}, \emph{BMI}, and \emph{Insulin} are the most important features, and the AdaBoost model has produced many decision stumps for these features (\autoref{fig:use_case2_summarization}(a)). What catches Amy's attention is the behavior of the stumps related to the \emph{Insulin} feature: From the six stumps related to this feature, three suggest that the prediction should be positive for diabetes, while the other three suggest the opposite. The \nth{3} decision stump, which has the highest weight value, appears to be divided between the two classes with a probability of about 50\% in the right subtree (\autoref{fig:use_case2_summarization}(b)). Due to this strange behavior, this stump is ruled out, and the \nth{11} decision stump provides the next most impactful rule in favor of the positive class prediction. After investigating the classification rules, Amy can easily override the problematic stump and adjust the model's behavior accordingly.

\textbf{Improving the surrogate model and communicating the results to domain experts.}
Amy wants to communicate her findings to the doctors. Since the currently active surrogate model has many stumps to analyze, which may overwhelm the doctors, she selects the one with complexity index \#6 instead (\autoref{fig:teaser}(a.1); \textbf{AT2}). This model has only eight stumps, has a fidelity of over 90\%. Interestingly, \textsc{DeforestVis} demonstrates that the next five surrogates use 2 decimal digits instead of 4 (gray dots being on top of black) for achieving higher fidelity. To ensure that she has learned the most from other pre-trained surrogate models with above 92\% fidelity, Amy checks for unique rules occurring after this AdaBoost model (i.e., having longer bars, see~\autoref{fig:teaser}(a.2); \textbf{AT3}) and selects the one from model \#14 with a moderate weighted probability value (red colored and dashed lines, see \autoref{fig:teaser}(a.2); \textbf{AT3}) because the remaining unique stumps have very low weighted probability (dark blue color). The decision stump suggests that, for \emph{Glucose}, the threshold should be set to $\approx$0.55 so that training samples below get classified as negative (\autoref{fig:teaser}(a.3)). This suggestion makes Amy think about the most impactful feature, \emph{Glucose}, which makes the model predict one class or the other, as suggested by the test samples in~\autoref{fig:teaser}(e) (\textbf{AT5}). Another finding is that \emph{Insulin} in this surrogate model is only positive, leading to fewer negative diabetes cases (fully green for all decisions). The selected \nth{6} decision stump (\autoref{fig:teaser}(c.1)) is much more in favor of the negative class with a very high threshold for proposing the positive class (\autoref{fig:teaser}(b.2)). With this knowledge, Amy decides to decrease the threshold for that stump to make the prediction more flexible (\autoref{fig:teaser}(c.2); \textbf{AT4}). The impact is visible in~\autoref{fig:teaser}(d) (\textbf{AT4}). Amy decides to present this finding to the experienced doctors to get their opinion about this manual threshold change and the behavior summary (\autoref{fig:teaser}(b.1)), as well as to verify the hypothesis that the updated surrogate model performs as they expect (or better than before) and potentially insert their knowledge into it.

\section{Evaluation} \label{sec:eval}
	We gathered more feedback on the  effectiveness of \textsc{\textsc{DeforestVis}} by conducting online semi-structured interview sessions with five experts (\textbf{E1}--\textbf{E5}), along the same procedure as in\,\cite{Ma2020Explaining,Xu2019EnsembleLens,Chatzimparmpas2022FeatureEnVi,Chatzimparmpas2023HardVis}. 

\subsection{Participants}

\textbf{E1} is an assistant professor with a PhD in mathematics and 7 years of experience with ML, currently developing ML models for reinforcement. \textbf{E2} is a senior researcher in a governmental research institute, working with applied ML projects, with a PhD in software engineering, and 6 years of experience with ML. \textbf{E3} is a data analyst in a large multinational company working with data engineering, a PhD in informatics, and 6 years of ML experience. \textbf{E4} is a data analyst and PhD candidate working with time-series data and anomaly detection with 5 years of ML experience. \textbf{E5} is a PhD candidate in deep learning with 5 years of experience in deploying ML models in a large multinational company. \textbf{E5} was the only one who reported a colorblindness issue (deuteranomaly), but mentioned having no problem perceiving correctly the colors used in \textsc{\textsc{DeforestVis}}. \textbf{All experts} were knowledgeable in data visualization and had encountered visual tools at some point during their professional careers.

\subsection{Methodology}

We conducted individual interview sessions online via Zoom using a large PC screen in full-screen mode, with the experts' participation being completely voluntary. \hl{Each interview session lasted about 1 hour and 30 minutes} and was structured as follows: (1) present the core research goals of \textsc{\textsc{DeforestVis}}, its analytical tasks (\autoref{sec:goals}) and workflow (\autoref{sec:overview}); (2) explain the functionality of every view and the steps taken to arrive at the results in~\autoref{sec:application}; and (3) interact with the tool on a newly-introduced data set (heart disease diagnosis~\cite{Dua2017}), similarly to the demo video accompanying this paper. The goal of phase (3) was not to accomplish a specific task, but rather to elucidate the interconnections between each view and to explore the potential capabilities of our system with this simplistic healthcare data set. In this formative evaluation, experts were asked to provide their opinions on the four aspects summarized in Sections~\ref{sec:intwork}--\ref{sec:intlim} by following a think-aloud protocol.

\subsection{Overview} 

The feedback we received was positive and supported the use of \textsc{DeforestVis} for surrogate modeling. \textbf{E4} highlighted the issue of adding more and more rules when using tree-based surrogates, making it rather quickly almost impossible to examine each one. A summarization of which trees are more interesting in the behavior model summarization and decisions comparison are great additions of \textsc{DeforestVis}, offering users further guidance (\textbf{RQ1}). \textbf{E1} and \textbf{E3} deemed our VA tool suitable for real-world applications and also educational purposes (the latter since we visually explain how AdaBoost works). \textbf{All experts} were impressed and expressed confidence in the advantages of using \textsc{DeforestVis}, especially praised for the transparency our VA tool offers from multiple levels (top-down, in-between, and bottom-up) and for various users (\textbf{RQ2}).

\subsection{Workflow} \label{sec:intwork}

\textbf{Somewhat intensive to learn but usable after training.} \textbf{E1} and \textbf{E4} praised the conceptual workflow of our tool, going from a broader view (their favorite panel is shown in \autoref{fig:teaser}(a)) to more fine-grained views. They both mentioned that the learning curve was steep. Interestingly, \textbf{E1} felt confident that he could understand the tool without a training session. \textbf{E4} suggested that some indicators could guide users on where to look first, but the training we provided was sufficient to make him understand how the tool works.

\textbf{Diverse workflow steps for different users.} \textbf{E2} and \textbf{E4} recommended having a model developer work together with a domain expert to enhance collaboration, which is an important aspect for visualization tools in general beyond \textsc{DeforestVis}. The top-down approach of selecting and tuning the surrogate model was more appropriate for model developers and data analysts (see \autoref{sec:case}); the bottom-up approach was found relevant for domain experts (i.e., starting from \autoref{fig:teaser}(e)). The rule-overriding related views (\autoref{fig:teaser}(c) and (d)) serve as the middle ground, enabling developers and experts to collaborate (\textbf{AT4}). \textbf{E4} suggested that domain experts could have better understood the overriding rule process if they had focused on feature-based modifications rather than decision stumps---a hypothesis that should be tested in the future. Still, the benefit of working with decision stumps is that it allows for micromanagement for each stump, which sometimes gets combined with other decision stumps (targeted toward data analysts and model developers). Here, \textbf{E4} pointed out that domain experts might need to change decision stumps based on their previous knowledge, that is, achievable by rule overriding and decisions comparison views in \textsc{DeforestVis}; and \textbf{E2} recommended that prior knowledge of AdaBoost might be required to explore the whole process (e.g., the background information provided in~\autoref{sec:back}).

\textbf{Extra verification loop.} Finally, \textbf{E2} proposed comparing the model-based extracted distributions against the actual data distribution to ensure that the model behaves as expected to reassure users to trust the results of the selected surrogate model. This topic is timely and requires further investigation by the visualization community (e.g., see Kale et al.~\cite{Kale2023EVM}).

\subsection{Visualization and coordination}

\textbf{Encouraging results about the novelty of \textsc{DeforestVis}.} \textbf{E4} stated that the interaction between performance and uniqueness (cf.~\autoref{fig:teaser}(a), top-right toggle) is powerful as it helps find decision stumps that are less influential and duplicated, showing they are uninteresting for users and guiding them to select a surrogate model (\textbf{AT1} and \textbf{AT2}). \cit{The model is simplified by reducing the number of one-level decision trees and rounding the threshold values to fewer decimals, which is incredible!}, said \textbf{E3} (\textbf{AT1}). \textbf{All experts} clicked on a decision tree in the dot plot with lines (\autoref{fig:teaser}(a)) to propagate to the following surrogate models, giving an idea of when this tree was re-weighted due to a duplicate decision stump, as mentioned by \textbf{E1}. \textbf{E1} followed up by saying: \cit{This view answers the question of how much complexity should be added before making a simple rule much more complex!} (\textbf{AT2}). As an analogy, \textsc{DeforestVis} could be viewed as facilitating users in inspecting diverse systematically-reduced dimensionalities, and effectively deciding the ``optimal'' one for the given problem~\cite{Jeong2023Dimensionality}.

\textbf{Improving visualizations.} \textbf{E1}, \textbf{E2}, and \textbf{E5} liked the fragmented bar chart in~\autoref{fig:teaser}(e) and found it useful for visualizing data and hypothesizing changes using a bottom-up approach (especially the $\Delta(W \times P)$ column; \textbf{AT5}). Additionally, \textbf{E5} stated that, for binary classification, ground truth covers the prediction, so the \emph{Pred} column is not necessary unless it is a multi-class problem. He suggested replacing the column with the predictions made by the targeted model to enable a direct comparison of whether the surrogate and target models agree or not. This idea led us to think about a future research opportunity where the input data are compared against the surrogate model's predictions except solely for the target model to improve directly the former by studying the misclassifications of the latter.

\textbf{Familiarity with the tool and coordination between views.} In the projection-based view, two potential drawbacks were found by \textbf{E3} and \textbf{E5} (\textbf{AT4}). The first issue is the abstract nature of the UMAP projection, which can make it hard to see prominent clusters of data points as the number of decision stumps increases (\textbf{E3}). A solution to this would be to highlight a specific training instance and see where it belongs to in all decision stumps at once. The second issue is the difficulty of keeping track of left \emph{vs} right subtrees, which can predict different classes in different decision trees (\textbf{E5}). This issue may improve with practice and familiarity with the data set, but also with the hovering functionality that can help explaining in which subtree an instance belongs to for every decision stump.

\subsection{Interaction}

\textbf{Interactivity improvements.} \textbf{E1} said that a lasso selection could be used in the UMAP view to find samples classified similarly in the same subtrees by decision stumps because analyzing samples individually in the training set is helpful but fails to directly explain sample clusters (\textbf{AT3} and \textbf{AT4}). However, he also said that the projection view gives insights into why samples are close and how confused they are. Also, trajectories (gray lines) are useful to show the global effect of local changes if clear clusters are formed by UMAP (see~\autoref{fig:teaser}(d); T4).

\textbf{Details on demand.} \textbf{E2} said that comparing stump-based weights is easy in pairs but by what exact amount is challenging, which could be improved by showing the precise value on hovering. These are all minor issues to be addressed in a future version of \textsc{DeforestVis}.

\textbf{Alternative color mapping.} Finally, \textbf{E3} suggested that the color gradient in~\autoref{fig:teaser}(a.2) could have been reversed from yellow to blue as yellow was perceived as stronger, but there is a design trade-off with consistency as the same color gradient is used in~\autoref{fig:teaser}(c.1), right-hand side (found well designed by them).

\subsection{Limitations identified by the experts} \label{sec:intlim}

\textbf{Scalability issues.} \textbf{E2} and \textbf{E5} were concerned about our tool's (1) \emph{scalability} \emph{vs} the \emph{number of features}; \textbf{E3} pointed the issue of (2) \emph{adding more training and testing samples}; \textbf{E4} pointed the issue of incrementally exploring (3) \emph{additional decision stumps}. For (1), \textbf{E2} and \textbf{E5} agreed with using AdaBoost as a surrogate model to fit target-model predictions as it automatically generates decision stumps only for important features. This is affected by the number of decision stumps of the selected surrogate model. For (2), \textbf{E3} proposed to aggregate similar training samples to scale the decision-comparison view and a bar chart similar to~\autoref{fig:teaser}(a.3) instead of the sample grids (\autoref{fig:teaser}(b.2)), and then doing the same for testing samples with binning test cases according to how they were predicted and how hard it is to force them to swap classes. \textbf{E5} suggested the progressive visual exploration of the data and features to overcome such issues. For (3), \textbf{E4} said that, with the surrogate model selection panel (\autoref{fig:teaser}(a)), users would typically select fewer decision stumps that are easily explorable, and worst case, explanations using decision stumps could be replaced by more segmented bar charts to show more features (also solving the scalability issue). 

Overall, \textsc{DeforestVis} has limited scalability concerning the instance and
feature count, that is, it can handle, conservatively, up to several thousands of instances having several tens of features. However, similar to other VA tools, \textsc{DeforestVis} does not aim to offer hundreds or thousands of individual
features to the user to construct an explanation. It thus falls in the group of tools that address so-called tabular data, which
consists of a relatively small number of features that have a clear meaning and identity that experts want to see
and reason about. The other group of tools address inherently high-dimensional data, e.g., with thousands of dimensions that
do not all have a clear meaning, and which one does not typically want to reason about explicitly, thus visualize
explicitly, as common in deep-learning contexts.

\textbf{Beyond binary classification.} \textbf{E1} noted that multi-class classification is perhaps the most limiting factor when using AdaBoost as a surrogate model (simple if-else decision stumps), but a binary one-vs-rest strategy could be used. Despite that, \textbf{E1} thought that the proposed visual designs are not bounded by this same restriction, especially if users focus on the segmented bar charts that summarize the predictive power and thresholds for each feature (\autoref{fig:teaser}(b.1)). However, in general, many related works focus only on binary classification problems according to Chatzimparmpas et al.~\cite{Chatzimparmpas2020The}, that is, the same category \textsc{DeforestVis} falls into.

\textbf{Held-out test set.} \textbf{E5} mentioned that there is a danger of bias if users tune the model to improve against the test cases presented in our tool. \textbf{E1} also mentioned that users should only use the bottom-up approach for explanation purposes and not for tampering with the training process. An external data set may be needed if our tool is used to tune the training of a surrogate model to fit well a test set, according to \textbf{E5}.

\section{Discussion} \label{sec:disc}
	\hl{In this section, we describe the intended target users and the general limitations of our VA tool beyond the ones identified by the experts.}

\subsection{Targeted users} \label{sec:users}

\textsc{DeforestVis} aims to assist model developers in understanding the behavior of their models in order to further optimize them, and also help domain experts to use simplified and tunable surrogate models for prediction instead of a more complex ML model. By visually examining surrogate models, developers can gain insights into how these surrogates represent the original model, which can help them improve both models. Visual examination of surrogate models can also help developers to identify influential features, detect interactions between features, and make other adjustments to improve the accuracy and reliability of the complex ML model, e.g., by informing future data collection cycles. For data analysts and domain experts, visualizations of surrogate models can provide an intuitive understanding of how these models behave and the relationships between the input variables and the output predictions. This can help to identify patterns, anomalies, and other important features in the data that may not be immediately apparent from numerical summaries or other types of analyses. Additionally, tree- and rule-based visualizations can assist data analysts in communicating their findings to domain experts in a clear and compelling way~\cite{Yuan2021An}. As highlighted in~\autoref{sec:eval}, the five ML experts who engaged in our interview sessions, each lasting 1 hour and 30 minutes, successfully comprehended the core principles and managed to use \textsc{DeforestVis}. \hl{An additional opportunity that emerges from this is the development of a simplified version of our tool, specifically designed for novice ML users with limited visualization literacy.}

\hl{Concretely, from our overall observations during the expert interviews (see~\autoref{sec:eval}), it is arguable that \textsc{DeforestVis} has a relatively steep learning curve---at least if one aims to master all its offered features. Indeed, our tool consists of five views (see~\autoref{fig:teaser} and~\autoref{fig:workflow-diagram}), each of them offering internally interactive sorting and/or selection mechanisms which, when activated, affect all the other linked views. Apart from the limitations exposed by our expert interviews, including the users’ suggestions for improvement (\autoref{sec:eval}), we see further possibilities to improve \textsc{DeforestVis} to reduce its learning curve:}

\begin{itemize}
\item \hl{Add \emph{explanatory tooltips} on the interactable elements in each view to tell users which are the available controls therein and what is their purpose. This should remind users what they can do within each view.}
\item \hl{Use a \emph{recommender system} that analyzes the last actions of the user and next recommends other views/controls in the tool that would naturally follow the given actions. For example, when a user has selected a surrogate model (\autoref{fig:teaser}, (a)), the tool can suggest, by means of a highlight, the examination of other surrogate models with unique decision stumps. This should guide users through a typical workflow based on their previous actions.}
\item \hl{Offer the use of \emph{presets} during a tool start-up dialog allowing setting problem-specific constraints. For example, users may never want to have a fidelity of less than 90\% of the targeted model or to explore more than a certain number of decision stumps. These constraints can next be used by \textsc{DeforestVis} to filter out, from all views, the data that does not fit them, leading to simpler, more targeted visualizations.}
\end{itemize}

\subsection{Limitations and potential extensions}

We next discuss several limitations of our proposal as well as potential improvements to alleviate these.

\noindent\textbf{Different fidelity measures.} Fidelity in eXplainable AI (XAI) refers to how well a surrogate model can emulate a complex, black-box model and aims to evaluate the surrogate's prediction ability as a whole~\cite{Velmurugan2021Evaluating}. A prevalent measure for global fidelity (also used by us) is the overall prediction accuracy of the surrogate model. Other global fidelity measures include the R-squared measure for regression tasks~\cite{Cameron1997An}. While intuitive and straightforward, such measures may fall short in cases with imbalanced data or varying misclassification costs. Measures can also be internal, \emph{e.g.}, assessing the surrogate's representation of the target model's decision boundaries~\cite{Velmurugan2021Evaluating}. However, such measures are relatively more complicated to compute and visualize. We next consider extending our approach to incorporate such more complex measures in a visually intuitive way.

\noindent\textbf{Beyond weak models.} Although the benefit of using decision stumps is their simplicity, this forces each model to be a weak predictor that may not capture all relations in the data. As each stump only considers one feature at a time, the interaction between features might be ignored, leading to less accurate predictions in situations where feature interactions are critical. Hence, the effectiveness of the explanations might be limited, as is the case with GAMs~\cite{Hastie1986Generalized} and neural additive models~\cite{Agarwal2021Neural} (a variant of GAMs for neural networks). Since the boosting procedure tries incrementally improving the model by focusing on instances misclassified in previous iterations, users can more easily intervene by adapting the individual stumps and observing their impact on the entire surrogate model~\cite{Wang2022Interpretability}. Ways to visualize more complex decision trees while the experts still wish to explore and modify them are yet to be found.

\noindent\textbf{Backpropagating the surrogate's changes to the target model.} While, in some cases, the surrogate model may completely \emph{replace} the target model (if accurate enough and since easier to explain and tune), in other cases one would like to incorporate the users' manual adjustments and insights into the target model. It is important to note that our current approach does not include this feedback mechanism. Rather, our emphasis has been on developing a more accessible \emph{alternative} to a complex model, that is, a surrogate model that domain experts can better comprehend and manipulate. Incorporating a feedback loop that can help refine the target model based on surrogate threshold adjustments is, to our knowledge, a generally unaddressed, but very important, challenge for future research.

\section{Conclusion} \label{sec:con}
	In this paper, we present \textsc{DeforestVis}, a VA tool that helps the analysis of black box, complex, ML models. We use decision stumps (one-level decision trees) from the AdaBoost algorithm, which are easily interpretable, to produce surrogate models that approximate the behavior of a target model. Our tool's linked views help users iteratively find a balance between complexity and fidelity while minimizing the precision (measured as number of decimals) used in the decision thresholds without sacrificing accuracy. Additionally, \textsc{DeforestVis} enables users to explore decision stumps in many ways, including their purity and impactfulness, and summarizes the behavior of the surrogate model by aggregating the predictive outcomes and power of all decision stumps related to each data feature. Users can override rules and compare them on local and global scales. Also, users can focus on particular test cases and use the extracted rules to understand why these cases got their particular classifications. We evaluated the applicability and effectiveness of \textsc{DeforestVis} using real-world data sets and by conducting expert interviews with five expert data analysts and model developers. Our results show the benefits of using \textsc{DeforestVis} for ML model analysis but also highlight potential limitations that we aim to address in future work.

\section*{Acknowledgements}
This work was partially supported through the ELLIIT environment for strategic research in Sweden.

\bibliographystyle{eg-alpha-doi}

\bibliography{DeforestVis}

\newcommand{\etalchar}[1]{$^{#1}$}
\begin{thebibliography}{\uppercase{NWWH19}}

\bibitem[AEK00]{Ankerst2000Towards}
\textsc{Ankerst M., Ester M., Kriegel H.-P.}:
\newblock Towards an effective cooperation of the user and the computer for
  classification.
\newblock In \emph{Proceedings of the Sixth ACM SIGKDD International Conference
  on Knowledge Discovery and Data Mining} (2000), KDD '00, ACM, pp.~179--188.
\newblock \href {https://doi.org/10.1145/347090.347124}
  {\path{doi:10.1145/347090.347124}}.

\bibitem[AF22]{Antweiler2022Visualizing}
\textsc{Antweiler D., Fuchs G.}:
\newblock Visualizing rule-based classifiers for clinical risk prognosis.
\newblock In \emph{Proceedings of the IEEE Visualization and Visual Analytics}
  (2022), VIS '22, IEEE, pp.~55--59.
\newblock \href {https://doi.org/10.1109/VIS54862.2022.00020}
  {\path{doi:10.1109/VIS54862.2022.00020}}.

\bibitem[AMF{\etalchar{*}}21]{Agarwal2021Neural}
\textsc{Agarwal R., Melnick L., Frosst N., Zhang X., Lengerich B., Caruana R.,
  Hinton G.~E.}:
\newblock Neural {A}dditive {M}odels: {I}nterpretable machine learning with
  neural nets.
\newblock In \emph{Advances in Neural Information Processing Systems} (2021),
  vol.~34, Curran Associates, Inc., pp.~4699--4711.

\bibitem[BKSS14]{Behrisch2014Feedback}
\textsc{Behrisch M., Korkmaz F., Shao L., Schreck T.}:
\newblock Feedback-driven interactive exploration of large multidimensional
  data supported by visual classifier.
\newblock In \emph{Proceedings of the IEEE Conference on Visual Analytics
  Science and Technology} (2014), VAST '14, pp.~43--52.
\newblock \href {https://doi.org/10.1109/VAST.2014.7042480}
  {\path{doi:10.1109/VAST.2014.7042480}}.

\bibitem[BN01]{Barlow2001Case}
\textsc{Barlow T., Neville P.}:
\newblock Case study: {V}isualization for decision tree analysis in data
  mining.
\newblock In \emph{Proceedings of the IEEE Symposium on Information
  Visualization,} (2001), INFOVIS '01, IEEE, pp.~149--152.
\newblock \href {https://doi.org/10.1109/INFVIS.2001.963292}
  {\path{doi:10.1109/INFVIS.2001.963292}}.

\bibitem[Bre01]{Breiman2001Random}
\textsc{Breiman L.}:
\newblock Random forests.
\newblock \emph{Machine Learning 45}, 1 (Oct. 2001), 5--32.
\newblock \href {https://doi.org/10.1023/A:1010933404324}
  {\path{doi:10.1023/A:1010933404324}}.

\bibitem[BvLH{\etalchar{*}}11]{Bremm2011Interactive}
\textsc{Bremm S., von Landesberger T., Heß M., Schreck T., Weil P., Hamacherk
  K.}:
\newblock Interactive visual comparison of multiple trees.
\newblock In \emph{Proceedings of the IEEE Conference on Visual Analytics
  Science and Technology} (2011), VAST '11, IEEE, pp.~31--40.
\newblock \href {https://doi.org/10.1109/VAST.2011.6102439}
  {\path{doi:10.1109/VAST.2011.6102439}}.

\bibitem[CB20]{Cao2020DRIL}
\textsc{Cao F., Brown E.~T.}:
\newblock {DRIL}: {D}escriptive rules by interactive learning.
\newblock In \emph{Proceedings of the IEEE Visualization Conference} (2020),
  VIS '20, pp.~256--260.
\newblock \href {https://doi.org/10.1109/VIS47514.2020.00058}
  {\path{doi:10.1109/VIS47514.2020.00058}}.

\bibitem[CD19]{Cavallo2019Clustrophile}
\textsc{Cavallo M., Demiralp C.}:
\newblock {Clustrophile 2}: {G}uided visual clustering analysis.
\newblock \emph{IEEE TVCG 25}, 1 (2019), 267--276.
\newblock \href {https://doi.org/10.1109/TVCG.2018.2864477}
  {\path{doi:10.1109/TVCG.2018.2864477}}.

\bibitem[CG16]{Chen2016XGBoost}
\textsc{Chen T., Guestrin C.}:
\newblock {XGBoost}: A scalable tree boosting system.
\newblock In \emph{Proceedings of the 22nd ACM SIGKDD International Conference
  on Knowledge Discovery and Data Mining} (2016), KDD~'16, ACM, pp.~785--794.
\newblock \href {https://doi.org/10.1145/2939672.2939785}
  {\path{doi:10.1145/2939672.2939785}}.

\bibitem[CLG{\etalchar{*}}15]{Caruana2015Intelligible}
\textsc{Caruana R., Lou Y., Gehrke J., Koch P., Sturm M., Elhadad N.}:
\newblock Intelligible models for healthcare: {P}redicting pneumonia risk and
  hospital 30-day readmission.
\newblock In \emph{Proceedings of the 21th ACM SIGKDD International Conference
  on Knowledge Discovery and Data Mining} (2015), KDD '15, ACM, p.~1721–1730.
\newblock \href {https://doi.org/10.1145/2783258.2788613}
  {\path{doi:10.1145/2783258.2788613}}.

\bibitem[CMJ{\etalchar{*}}20]{Chatzimparmpas2020The}
\textsc{Chatzimparmpas A., Martins R.~M., Jusufi I., Kucher K., Rossi F.,
  Kerren A.}:
\newblock The state of the art in enhancing trust in machine learning models
  with the use of visualizations.
\newblock \emph{Computer Graphics Forum 39}, 3 (June 2020), 713--756.
\newblock \href {https://doi.org/10.1111/cgf.14034}
  {\path{doi:10.1111/cgf.14034}}.

\bibitem[CMJK20]{Chatzimparmpas2020A}
\textsc{Chatzimparmpas A., Martins R.~M., Jusufi I., Kerren A.}:
\newblock A survey of surveys on the use of visualization for interpreting
  machine learning models.
\newblock \emph{Information Visualization 19}, 3 (July 2020), 207--233.
\newblock \href {https://doi.org/10.1177/1473871620904671}
  {\path{doi:10.1177/1473871620904671}}.

\bibitem[CMK20]{Chatzimparmpas2020t}
\textsc{Chatzimparmpas A., Martins R.~M., Kerren A.}:
\newblock {t-viSNE}: Interactive assessment and interpretation of {t-SNE}
  projections.
\newblock \emph{IEEE TVCG 26}, 8 (Aug. 2020), 2696--2714.
\newblock \href {https://doi.org/10.1109/TVCG.2020.2986996}
  {\path{doi:10.1109/TVCG.2020.2986996}}.

\bibitem[CMK23]{Chatzimparmpas2023VisRuler}
\textsc{Chatzimparmpas A., Martins R.~M., Kerren A.}:
\newblock Vis{R}uler: Visual analytics for extracting decision rules from
  bagged and boosted decision trees.
\newblock \emph{Information Visualization 22}, 2 (2023), 115--139.
\newblock \href {https://doi.org/10.1177/14738716221142005}
  {\path{doi:10.1177/14738716221142005}}.

\bibitem[CMKK21]{Chatzimparmpas2021StackGenVis}
\textsc{Chatzimparmpas A., Martins R.~M., Kucher K., Kerren A.}:
\newblock {StackGenVis}: {A}lignment of data, algorithms, and models for
  stacking ensemble learning using performance metrics.
\newblock \emph{IEEE TVCG 27}, 2 (2021), 1547--1557.
\newblock \href {https://doi.org/10.1109/TVCG.2020.3030352}
  {\path{doi:10.1109/TVCG.2020.3030352}}.

\bibitem[CMKK22]{Chatzimparmpas2022FeatureEnVi}
\textsc{Chatzimparmpas A., Martins R.~M., Kucher K., Kerren A.}:
\newblock Feature{E}n{V}i: Visual analytics for feature engineering using
  stepwise selection and semi-automatic extraction approaches.
\newblock \emph{IEEE TVCG 28}, 4 (2022), 1773--1791.
\newblock \href {https://doi.org/10.1109/TVCG.2022.3141040}
  {\path{doi:10.1109/TVCG.2022.3141040}}.

\bibitem[CPK23]{Chatzimparmpas2023HardVis}
\textsc{Chatzimparmpas A., Paulovich F.~V., Kerren A.}:
\newblock Hard{V}is: Visual analytics to handle instance hardness using
  undersampling and oversampling techniques.
\newblock \emph{Computer Graphics Forum 42}, 1 (2023), 135--154.
\newblock \href {https://doi.org/https://doi.org/10.1111/cgf.14726}
  {\path{doi:https://doi.org/10.1111/cgf.14726}}.

\bibitem[CvW22]{Collaris2022StrategyAtlas}
\textsc{Collaris D., van Wijk J.}:
\newblock {StrategyAtlas}: {S}trategy analysis for machine learning
  interpretability.
\newblock \emph{IEEE TVCG} (2022), 1--13.
\newblock To appear.
\newblock \href {https://doi.org/10.1109/TVCG.2022.3146806}
  {\path{doi:10.1109/TVCG.2022.3146806}}.

\bibitem[CW97]{Cameron1997An}
\textsc{{Colin Cameron} A., Windmeijer F.~A.}:
\newblock An {R}-squared measure of goodness of fit for some common nonlinear
  regression models.
\newblock \emph{Journal of Econometrics 77}, 2 (1997), 329--342.
\newblock \href {https://doi.org/10.1016/S0304-4076(96)01818-0}
  {\path{doi:10.1016/S0304-4076(96)01818-0}}.

\bibitem[D311]{D3}
{D3} --- {D}ata-driven documents, 2011.
\newblock Accessed January 24, 2024.
\newblock URL: \url{https://d3js.org/}.

\bibitem[DB21]{Agus2021RISSAD}
\textsc{Deng J., Brown E.~T.}:
\newblock {RISSAD}: {R}ule-based interactive semi-supervised anomaly detection.
\newblock In \emph{Proceedings of the EuroVis 2021 -- Short Papers} (2021), The
  Eurographics Association.
\newblock \href {https://doi.org/10.2312/evs.20211050}
  {\path{doi:10.2312/evs.20211050}}.

\bibitem[DCB19]{Castro2019Surrogate}
\textsc{Di~Castro F., Bertini E.}:
\newblock Surrogate decision tree visualization.
\newblock In \emph{Proceedings of the CEUR Workshop} (2019), vol.~2327,
  CEUR-WS.

\bibitem[Def23]{DeforestVisCode}
{DeforestVis Code}, 2023.
\newblock Accessed January 24, 2024.
\newblock URL: \url{https://github.com/angeloschatzimparmpas/DeforestVis}.

\bibitem[DG17]{Dua2017}
\textsc{Dua D., Graff C.}:
\newblock {UCI} machine learning repository, 2017.
\newblock Accessed January 24, 2024.
\newblock URL: \url{http://archive.ics.uci.edu/ml}.

\bibitem[Do07]{Do2007Towards}
\textsc{Do T.-N.}:
\newblock Towards simple, easy to understand, an interactive decision tree
  algorithm.
\newblock \emph{College Information Technology Can tho University, Can Tho,
  Vietnam, Technology Report} (2007), 06--01.

\bibitem[EAM14]{Eisemann2014A}
\textsc{Eisemann M., Albuquerque G., Magnor M.}:
\newblock A nested hierarchy of localized scatterplots.
\newblock In \emph{Proceedings of the 27th SIBGRAPI Conference on Graphics,
  Patterns and Images} (2014), pp.~80--86.
\newblock \href {https://doi.org/10.1109/SIBGRAPI.2014.14}
  {\path{doi:10.1109/SIBGRAPI.2014.14}}.

\bibitem[EAMS19]{Elshawi2019Interpretability}
\textsc{Elshawi R., Al-Mallah M.~H., Sakr S.}:
\newblock On the interpretability of machine learning-based model for
  predicting hypertension.
\newblock \emph{BMC Medical Informatics and Decision Making 19}, 1 (2019),
  1--32.

\bibitem[EF10]{Elmqvist2010Hierarchical}
\textsc{Elmqvist N., Fekete J.-D.}:
\newblock Hierarchical aggregation for information isualization: Overview,
  techniques, and design guidelines.
\newblock \emph{IEEE Transactions on Visualization and Computer Graphics 16}, 3
  (2010), 439--454.
\newblock \href {https://doi.org/10.1109/TVCG.2009.84}
  {\path{doi:10.1109/TVCG.2009.84}}.

\bibitem[EMJ{\etalchar{*}}22]{Eirich2022RfX}
\textsc{Eirich J., Münch M., Jäckle D., Sedlmair M., Bonart J., Schreck T.}:
\newblock {RfX}: A design study for the interactive exploration of a random
  forest to enhance testing procedures for electrical engines.
\newblock \emph{Computer Graphics Forum 41}, 6 (2022), 302--315.
\newblock \href {https://doi.org/10.1111/cgf.14452}
  {\path{doi:10.1111/cgf.14452}}.

\bibitem[Fla10]{Flask}
{Flask} --- {A} micro web framework written in {Python}, 2010.
\newblock Accessed January 24, 2024.
\newblock URL: \url{https://flask.palletsprojects.com/}.

\bibitem[Fri01]{Friedman2001Greedy}
\textsc{Friedman J.~H.}:
\newblock Greedy function approximation: {A} gradient boosting machine.
\newblock \emph{Annals of Statistics} (2001), 1189--1232.

\bibitem[FSA99]{Freund1999A}
\textsc{Freund Y., Schapire R., Abe N.}:
\newblock A short introduction to boosting.
\newblock \emph{Journal of Japanese Society for Artificial Intelligence 14}, 5
  (Sept. 1999), 771--780.

\bibitem[FW98]{Frank1998Generating}
\textsc{Frank E., Witten I.~H.}:
\newblock Generating accurate rule sets without global optimization.
\newblock In \emph{Proceedings of the Fifteenth International Conference on
  Machine Learning} (1998), ICML '98, Morgan Kaufmann Publishers Inc.,
  p.~144–151.

\bibitem[GGPPS13]{Gomez2013Visualizing}
\textsc{Guerra-Gómez J., Pack M.~L., Plaisant C., Shneiderman B.}:
\newblock Visualizing change over time using dynamic hierarchies:
  {TreeVersity2} and the {StemView}.
\newblock \emph{IEEE TVCG 19}, 12 (2013), 2566--2575.
\newblock \href {https://doi.org/10.1109/TVCG.2013.231}
  {\path{doi:10.1109/TVCG.2013.231}}.

\bibitem[GPTM10]{Genuer2010Variable}
\textsc{Genuer R., Poggi J.-M., Tuleau-Malot C.}:
\newblock Variable selection using random forests.
\newblock \emph{Pattern Recognition Letters 31}, 14 (2010), 2225--2236.

\bibitem[HC00]{Han2000RuleViz}
\textsc{Han J., Cercone N.}:
\newblock {RuleViz}: {A} model for visualizing knowledge discovery process.
\newblock In \emph{Proceedings of the sixth ACM SIGKDD international conference
  on Knowledge discovery and data mining} (2000), KDD '00, ACM, p.~244–253.
\newblock \href {https://doi.org/10.1145/347090.347139}
  {\path{doi:10.1145/347090.347139}}.

\bibitem[HHC{\etalchar{*}}19]{Hohman2019Gamut}
\textsc{Hohman F., Head A., Caruana R., DeLine R., Drucker S.~M.}:
\newblock Gamut: A design probe to understand how data scientists understand
  machine learning models.
\newblock In \emph{Proceedings of the SIGCHI Conference on Human Factors in
  Computing Systems} (2019), ACM.
\newblock \href {https://doi.org/10.1145/3290605.3300809}
  {\path{doi:10.1145/3290605.3300809}}.

\bibitem[HKPC19]{Hohman2019Visual}
\textsc{Hohman F., Kahng M., Pienta R., Chau D.~H.}:
\newblock Visual analytics in deep learning: An interrogative survey for the
  next frontiers.
\newblock \emph{IEEE TVCG 25}, 8 (2019), 2674--2693.
\newblock \href {https://doi.org/10.1109/TVCG.2018.2843369}
  {\path{doi:10.1109/TVCG.2018.2843369}}.

\bibitem[HLLW19]{Huang2019GBRTVis}
\textsc{Huang Y., Liu Y., Li C., Wang C.}:
\newblock {GBRTVis}: {O}nline analysis of gradient boosting regression tree.
\newblock \emph{Journal of Visualization 22}, 1 (Feb. 2019), 125–140.
\newblock \href {https://doi.org/10.1007/s12650-018-0514-2}
  {\path{doi:10.1007/s12650-018-0514-2}}.

\bibitem[HMJE{\etalchar{*}}19]{Hilasaca2019Overlap}
\textsc{Hilasaca G.~M., Marc{\'\i}lio-Jr W.~E., Eler D.~M., Martins R.~M.,
  Paulovich F.~V.}:
\newblock Overlap removal of dimensionality reduction scatterplot layouts.
\newblock \emph{ArXiv e-prints 1903.06262} (2019).
\newblock \href {http://arxiv.org/abs/1903.06262} {\path{arXiv:1903.06262}}.

\bibitem[HT86]{Hastie1986Generalized}
\textsc{Hastie T., Tibshirani R.}:
\newblock Generalized additive models.
\newblock \emph{Statistical Science 1}, 3 (1986), 297 -- 310.
\newblock URL: \url{https://doi.org/10.1214/ss/1177013604}, \href
  {https://doi.org/10.1214/ss/1177013604} {\path{doi:10.1214/ss/1177013604}}.

\bibitem[IL92]{Iba1992Induction}
\textsc{Iba W., Langley P.}:
\newblock Induction of one-level decision trees.
\newblock In \emph{Machine Learning Proceedings 1992}. Morgan Kaufmann, San
  Francisco (CA), 1992, pp.~233--240.
\newblock \href {https://doi.org/10.1016/B978-1-55860-247-2.50035-8}
  {\path{doi:10.1016/B978-1-55860-247-2.50035-8}}.

\bibitem[JJLJ23]{Jeong2023Dimensionality}
\textsc{Jeong H., Jeong H.-o., Lee S., Jeong W.-K.}:
\newblock Dimensionality explorer for single-cell analysis.
\newblock In \emph{2023 IEEE 16th Pacific Visualization Symposium (PacificVis)}
  (2023), pp.~51--60.
\newblock \href {https://doi.org/10.1109/PacificVis56936.2023.00013}
  {\path{doi:10.1109/PacificVis56936.2023.00013}}.

\bibitem[JLL{\etalchar{*}}20]{Jia2020Visualizing}
\textsc{Jia S., Lin P., Li Z., Zhang J., Liu S.}:
\newblock Visualizing surrogate decision trees of convolutional neural
  networks.
\newblock \emph{Journal of Visualization 23}, 1 (2020), 141--156.
\newblock \href {https://doi.org/10.1007/s12650-019-00607-z}
  {\path{doi:10.1007/s12650-019-00607-z}}.

\bibitem[KGQ{\etalchar{*}}24]{Kale2023EVM}
\textsc{Kale A., Guo Z., Qiao X.~L., Heer J., Hullman J.}:
\newblock {EVM}: {I}ncorporating model checking into exploratory visual
  analysis, 2024.
\newblock To appear.

\bibitem[LJC16]{Lee2016An}
\textsc{Lee T., Johnson J., Cheng S.}:
\newblock An interactive machine learning framework, 2016.
\newblock \href {http://arxiv.org/abs/1610.05463} {\path{arXiv:1610.05463}}.

\bibitem[LL17]{Lundberg2017A}
\textsc{Lundberg S.~M., Lee S.-I.}:
\newblock A unified approach to interpreting model predictions.
\newblock In \emph{Proceedings of the Advances in Neural Information Processing
  Systems} (2017), vol.~30, Curran Associates, Inc.

\bibitem[LXL{\etalchar{*}}18]{Liu2018Visual}
\textsc{Liu S., Xiao J., Liu J., Wang X., Wu J., Zhu J.}:
\newblock Visual diagnosis of tree boosting methods.
\newblock \emph{IEEE TVCG 24}, 1 (Jan. 2018), 163--173.
\newblock \href {https://doi.org/10.1109/TVCG.2017.2744378}
  {\path{doi:10.1109/TVCG.2017.2744378}}.

\bibitem[MGT{\etalchar{*}}03]{Munzner2003TreeJuxtaposer}
\textsc{Munzner T., Guimbreti\`{e}re F., Tasiran S., Zhang L., Zhou Y.}:
\newblock {TreeJuxtaposer}: {S}calable tree comparison using focus+context with
  guaranteed visibility.
\newblock \emph{ACM Transactions on Graphics 22}, 3 (July 2003), 453–462.
\newblock \href {https://doi.org/10.1145/882262.882291}
  {\path{doi:10.1145/882262.882291}}.

\bibitem[MHM18]{McInnes2018UMAP}
\textsc{{McInnes} L., {Healy} J., {Melville} J.}:
\newblock {UMAP}: Uniform manifold approximation and projection for dimension
  reduction.
\newblock \emph{ArXiv e-prints 1802.03426} (Feb. 2018).
\newblock \href {http://arxiv.org/abs/1802.03426} {\path{arXiv:1802.03426}}.

\bibitem[MJEP{\etalchar{*}}21]{Marcilio2021ExplorerTree}
\textsc{Marcílio-Jr W.~E., Eler D.~M., Paulovich F.~V., Rodrigues-Jr J.~F.,
  Artero A.~O.}:
\newblock {ExplorerTree}: {A} focus+context exploration approach for {2D}
  embeddings.
\newblock \emph{Big Data Research 25} (2021), 100239.
\newblock \href {https://doi.org/10.1016/j.bdr.2021.100239}
  {\path{doi:10.1016/j.bdr.2021.100239}}.

\bibitem[MLMP18]{Muhlbacher2018TreePOD}
\textsc{Mühlbacher T., Linhardt L., Möller T., Piringer H.}:
\newblock Tree{POD}: Sensitivity-aware selection of pareto-optimal decision
  trees.
\newblock \emph{IEEE TVCG 24}, 1 (2018), 174--183.
\newblock \href {https://doi.org/10.1109/TVCG.2017.2745158}
  {\path{doi:10.1109/TVCG.2017.2745158}}.

\bibitem[Mol20]{Molnar2020Interpretable}
\textsc{Molnar C.}:
\newblock \emph{Interpretable Machine Learning}.
\newblock Independently Published, 2020.

\bibitem[MQB19]{Ming2019RuleMatrix}
\textsc{Ming Y., Qu H., Bertini E.}:
\newblock {RuleMatrix}: {V}isualizing and understanding classifiers with rules.
\newblock \emph{IEEE TVCG 25}, 1 (2019), 342--352.
\newblock \href {https://doi.org/10.1109/TVCG.2018.2864812}
  {\path{doi:10.1109/TVCG.2018.2864812}}.

\bibitem[Mun09]{Munzner2009A}
\textsc{Munzner T.}:
\newblock A nested model for visualization design and validation.
\newblock \emph{IEEE TVCG 15}, 6 (2009), 921--928.
\newblock \href {https://doi.org/10.1109/TVCG.2009.111}
  {\path{doi:10.1109/TVCG.2009.111}}.

\bibitem[Mun14]{Munzner2014visualization}
\textsc{Munzner T.}:
\newblock \emph{Visualization analysis and design}.
\newblock CRC press, 2014.

\bibitem[MXLM20]{Ma2020Explaining}
\textsc{Ma Y., Xie T., Li J., Maciejewski R.}:
\newblock Explaining vulnerabilities to adversarial machine learning through
  visual analytics.
\newblock \emph{IEEE TVCG 26}, 1 (Jan. 2020), 1075--1085.
\newblock \href {https://doi.org/10.1109/TVCG.2019.2934631}
  {\path{doi:10.1109/TVCG.2019.2934631}}.

\bibitem[NHS00]{Nguyen2000A}
\textsc{Nguyen T., Ho T., Shimodaira H.}:
\newblock A visualization tool for interactive learning of large decision
  trees.
\newblock In \emph{Proceedings of the 12th IEEE Internationals Conference on
  Tools with Artificial Intelligence} (2000), ICTAI '00, IEEE, pp.~28--35.
\newblock \href {https://doi.org/10.1109/TAI.2000.889842}
  {\path{doi:10.1109/TAI.2000.889842}}.

\bibitem[NJKC19]{Nori2019InterpretML}
\textsc{Nori H., Jenkins S., Koch P., Caruana R.}:
\newblock {InterpretML}: A unified framework for machine learning
  interpretability.
\newblock \emph{ArXiv e-prints 1909.09223} (Sep. 2019).

\bibitem[NP21]{Neto2021Explainable}
\textsc{Neto M.~P., Paulovich F.~V.}:
\newblock {Explainable Matrix} - {V}isualization for global and local
  interpretability of random forest classification ensembles.
\newblock \emph{IEEE TVCG 27}, 2 (2021), 1427--1437.
\newblock \href {https://doi.org/10.1109/TVCG.2020.3030354}
  {\path{doi:10.1109/TVCG.2020.3030354}}.

\bibitem[NP22]{Neto2021Multivariate}
\textsc{Neto M.~P., Paulovich F.~V.}:
\newblock Multivariate data explanation by {J}umping {E}merging {P}atterns
  visualization.
\newblock \emph{IEEE TVCG} (2022), 1--16.
\newblock To appear.
\newblock \href {https://doi.org/10.1109/TVCG.2022.3223529}
  {\path{doi:10.1109/TVCG.2022.3223529}}.

\bibitem[NWWH19]{Nsch2019Colorful}
\textsc{Nsch R.~H., Wiesner P., Wendler S., Hellwich O.}:
\newblock Colorful {T}rees: Visualizing random forests for analysis and
  interpretation.
\newblock In \emph{Proceedings of the IEEE Winter Conference on Applications of
  Computer Vision} (2019), WACV '19, IEEE, pp.~294--302.
\newblock \href {https://doi.org/10.1109/WACV.2019.00037}
  {\path{doi:10.1109/WACV.2019.00037}}.

\bibitem[plo10]{plotly}
{Plotly} --- {JavaScript} open source graphing library, 2010.
\newblock Accessed January 24, 2024.
\newblock URL: \url{https://plotly.com}.

\bibitem[PNWG17]{Phillips2017FFTrees}
\textsc{Phillips N.~D., Neth H., Woike J.~K., Gaissmaier W.}:
\newblock {FFTrees}: {A} toolbox to create, visualize, and evaluate
  fast-and-frugal decision trees.
\newblock \emph{Judgment and Decision making 12}, 4 (2017), 344--368.

\bibitem[PSMD14]{Padua2014Interactive}
\textsc{Padua L., Schulze H., Matković K., Delrieux C.}:
\newblock Interactive exploration of parameter space in data mining:
  {C}omprehending the predictive quality of large decision tree collections.
\newblock \emph{Computers \& Graphics 41} (2014), 99--113.
\newblock \href {https://doi.org/10.1016/j.cag.2014.02.004}
  {\path{doi:10.1016/j.cag.2014.02.004}}.

\bibitem[PVG{\etalchar{*}}11]{Pedregosa2011Scikit}
\textsc{Pedregosa F., Varoquaux G., Gramfort A., Michel V., Thirion B., Grisel
  O., Blondel M., Prettenhofer P., Weiss R., Dubourg V., Vanderplas J., Passos
  A., Cournapeau D., Brucher M., Perrot M., Duchesnay E.}:
\newblock {Scikit-Learn}: Machine learning in {P}ython.
\newblock \emph{Journal of Machine Learning Research 12} (Nov. 2011),
  2825--2830.
\newblock \href {https://doi.org/10.5555/1953048.2078195}
  {\path{doi:10.5555/1953048.2078195}}.

\bibitem[RSG16]{Ribeiro2016Why}
\textsc{Ribeiro M.~T., Singh S., Guestrin C.}:
\newblock ``{W}hy should {I} trust you?'': Explaining the predictions of any
  classifier.
\newblock In \emph{Proceedings of the 22nd ACM SIGKDD International Conference
  on Knowledge Discovery and Data Mining} (2016), KDD~'16, ACM, pp.~1135--1144.
\newblock \href {https://doi.org/10.1145/2939672.2939778}
  {\path{doi:10.1145/2939672.2939778}}.

\bibitem[RSG18]{Ribeiro2018Anchors}
\textsc{Ribeiro M.~T., Singh S., Guestrin C.}:
\newblock Anchors: High-precision model-agnostic explanations.
\newblock In \emph{Proceedings of the AAAI Conference on Artificial
  Intelligence} (Apr. 2018), vol.~32.
\newblock \href {https://doi.org/10.1609/aaai.v32i1.11491}
  {\path{doi:10.1609/aaai.v32i1.11491}}.

\bibitem[Sch99]{Schapire1999A}
\textsc{Schapire R.~E.}:
\newblock A brief introduction to boosting.
\newblock In \emph{Proceedings of the 16th International Joint Conference on
  Artificial Intelligence - Volume 2} (1999), IJCAI'99, Morgan Kaufmann
  Publishers Inc., p.~1401–1406.

\bibitem[SCS04]{Hongzhi2004Multiple}
\textsc{Song H., Curran E.~P., Sterritt R.}:
\newblock Multiple foci visualisation of large hierarchies with {FlexTree}.
\newblock \emph{Information Visualization 3}, 1 (2004), 19--35.
\newblock \href {https://doi.org/10.1057/palgrave.ivs.9500065}
  {\path{doi:10.1057/palgrave.ivs.9500065}}.

\bibitem[SED{\etalchar{*}}88]{Smith1988Using}
\textsc{{Smith} J., {Everhart} J., {Dickson} W., {Knowler} W., {Johannes} R.}:
\newblock Using the {ADAP} learning algorithm to forecast the onset of diabetes
  mellitus.
\newblock In \emph{Proceedings of the Annual Symposium Computer Application in
  Medical Care} (1988), American Medical Informatics Association, pp.~261--265.

\bibitem[SFK08]{Sobester2008Engineering}
\textsc{Sobester A., Forrester A., Keane A.}:
\newblock \emph{Engineering design via surrogate modelling: {A} practical
  guide}.
\newblock John Wiley \& Sons, 2008.

\bibitem[SL91]{Safavian1991A}
\textsc{Safavian S., Landgrebe D.}:
\newblock A survey of decision tree classifier methodology.
\newblock \emph{IEEE Transactions on Systems, Man, and Cybernetics 21}, 3
  (1991), 660--674.
\newblock \href {https://doi.org/10.1109/21.97458}
  {\path{doi:10.1109/21.97458}}.

\bibitem[ST01]{Sato2001Rule}
\textsc{Sato M., Tsukimoto H.}:
\newblock Rule extraction from neural networks via decision tree induction.
\newblock In \emph{Proceedings of the International Joint Conference on Neural
  Networks} (2001), vol.~3 of \emph{IJCNN '01}, pp.~1870--1875 vol.3.
\newblock \href {https://doi.org/10.1109/IJCNN.2001.938448}
  {\path{doi:10.1109/IJCNN.2001.938448}}.

\bibitem[TDB21]{Thomas2021FacetRules}
\textsc{Thomas L.~V., Deng J., Brown E.~T.}:
\newblock Facet{R}ules: Discovering and describing related groups.
\newblock In \emph{Proceedings of the IEEE Workshop on Machine Learning from
  User Interactions} (2021), MLUI '21, pp.~21--26.
\newblock \href {https://doi.org/10.1109/MLUI54255.2021.00008}
  {\path{doi:10.1109/MLUI54255.2021.00008}}.

\bibitem[TKC17]{Tam2017An}
\textsc{Tam G. K.~L., Kothari V., Chen M.}:
\newblock An analysis of machine- and human-analytics in classification.
\newblock \emph{IEEE TVCG 23}, 1 (2017), 71--80.
\newblock \href {https://doi.org/10.1109/TVCG.2016.2598829}
  {\path{doi:10.1109/TVCG.2016.2598829}}.

\bibitem[TM03a]{Teoh2003PaintingClass}
\textsc{Teoh S.~T., Ma K.-L.}:
\newblock {PaintingClass}: {I}nteractive construction, visualization and
  exploration of decision trees.
\newblock In \emph{Proceedings of the ninth ACM SIGKDD international conference
  on Knowledge discovery and data mining} (2003), KDD '03, ACM, p.~667–672.
\newblock \href {https://doi.org/10.1145/956750.956837}
  {\path{doi:10.1145/956750.956837}}.

\bibitem[TM03b]{Teoh2003Starclass}
\textsc{Teoh S.~T., Ma K.-L.}:
\newblock {StarClass}: {I}nteractive visual classification using star
  coordinates.
\newblock In \emph{Proceedings of the 2003 SIAM International Conference on
  Data Mining} (2003), SIAM, pp.~178--185.
\newblock \href {https://doi.org/10.1137/1.9781611972733.16}
  {\path{doi:10.1137/1.9781611972733.16}}.

\bibitem[VCP22]{Varu2022ARMatrix}
\textsc{Varu R., Christino L., Paulovich F.~V.}:
\newblock {ARMatrix}: An interactive item-to-rule matrix for association rules
  visual analytics.
\newblock \emph{Electronics 11}, 9 (2022).
\newblock \href {https://doi.org/10.3390/electronics11091344}
  {\path{doi:10.3390/electronics11091344}}.

\bibitem[vdEvW11]{Elzen2011BaobabView}
\textsc{van~den Elzen S., van Wijk J.~J.}:
\newblock {BaobabView}: {I}nteractive construction and analysis of decision
  trees.
\newblock In \emph{Proceedings of the IEEE Conference on Visual Analytics
  Science and Technology} (2011), VAST '11, IEEE, pp.~151--160.
\newblock \href {https://doi.org/10.1109/VAST.2011.6102453}
  {\path{doi:10.1109/VAST.2011.6102453}}.

\bibitem[vdMH08]{vanDerMaaten2008Visualizing}
\textsc{van~der Maaten L., Hinton G.}:
\newblock Visualizing data using {t-SNE}.
\newblock \emph{Journal of Machine Learning Research 9} (2008), 2579--2605.

\bibitem[VOMS21]{Velmurugan2021Evaluating}
\textsc{Velmurugan M., Ouyang C., Moreira C., Sindhgatta R.}:
\newblock Evaluating fidelity of explainable methods for predictive process
  analytics.
\newblock In \emph{Proceedings of the International Conference on Advanced
  Information Systems Engineering} (2021), Springer, pp.~64--72.

\bibitem[vue14]{vuejs}
{Vue.js} --- {T}he progressive {JavaScript} framework, 2014.
\newblock Accessed January 24, 2024.
\newblock URL: \url{https://vuejs.org/}.

\bibitem[WFH{\etalchar{*}}01]{Ware2001Interactive}
\textsc{Ware M., Frank E., Holmes G., Hall M., Witten I.~H.}:
\newblock Interactive machine learning: {L}etting users build classifiers.
\newblock \emph{International Journal of Human-Computer Studies 55}, 3 (2001),
  281--292.
\newblock \href {https://doi.org/10.1006/ijhc.2001.0499}
  {\path{doi:10.1006/ijhc.2001.0499}}.

\bibitem[WKN{\etalchar{*}}22]{Wang2022Interpretability}
\textsc{Wang Z.~J., Kale A., Nori H., Stella P., Nunnally M.~E., Chau D.~H.,
  Vorvoreanu M., Wortman~Vaughan J., Caruana R.}:
\newblock Interpretability, then what? {E}diting machine learning models to
  reflect human knowledge and values.
\newblock In \emph{Proceedings of the 28th ACM SIGKDD Conference on Knowledge
  Discovery and Data Mining} (New York, NY, USA, 2022), KDD '22, Association
  for Computing Machinery, p.~4132–4142.
\newblock \href {https://doi.org/10.1145/3534678.3539074}
  {\path{doi:10.1145/3534678.3539074}}.

\bibitem[Wol92]{Wolpert1992Stacked}
\textsc{Wolpert D.~H.}:
\newblock Stacked generalization.
\newblock \emph{Neural Networks 5}, 2 (1992), 241--259.
\newblock \href {https://doi.org/10.1016/S0893-6080(05)80023-1}
  {\path{doi:10.1016/S0893-6080(05)80023-1}}.

\bibitem[WZWY21]{Wang2021Investigating}
\textsc{Wang J., Zhang W., Wang L., Yang H.}:
\newblock Investigating the evolution of tree boosting models with visual
  analytics.
\newblock In \emph{Proceedings of the 14th IEEE Pacific Visualization
  Symposium} (2021), PacificVis '21, IEEE, pp.~186--195.
\newblock \href {https://doi.org/10.1109/PacificVis52677.2021.00032}
  {\path{doi:10.1109/PacificVis52677.2021.00032}}.

\bibitem[XCC{\etalchar{*}}21]{Xia2021GBMVis}
\textsc{Xia Y., Cheng K., Cheng Z., Rao Y., Pu J.}:
\newblock {GBMVis}: Visual analytics for interpreting gradient boosting
  machine.
\newblock In \emph{Proceedings of the Cooperative Design, Visualization, and
  Engineering: 18th International Conference} (2021), CDVE '21, Springer,
  pp.~63--72.
\newblock \href {https://doi.org/10.1007/978-3-030-88207-5_7}
  {\path{doi:10.1007/978-3-030-88207-5_7}}.

\bibitem[XXM{\etalchar{*}}19]{Xu2019EnsembleLens}
\textsc{Xu K., Xia M., Mu X., Wang Y., Cao N.}:
\newblock {EnsembleLens}: Ensemble-based visual exploration of anomaly
  detection algorithms with multidimensional data.
\newblock \emph{IEEE TVCG 25}, 1 (Jan. 2019), 109--119.
\newblock \href {https://doi.org/10.1109/TVCG.2018.2864825}
  {\path{doi:10.1109/TVCG.2018.2864825}}.

\bibitem[YBOB22]{Yuan2022Visual}
\textsc{Yuan J., Barr B., Overton K., Bertini E.}:
\newblock Visual exploration of machine learning model behavior with
  hierarchical surrogate rule sets.
\newblock \emph{IEEE TVCG} (2022), 1--18.
\newblock To appear.
\newblock \href {https://doi.org/10.1109/TVCG.2022.3219232}
  {\path{doi:10.1109/TVCG.2022.3219232}}.

\bibitem[YCB{\etalchar{*}}22]{Yuan2022SUBPLEX}
\textsc{Yuan J., Chan G. Y.-Y., Barr B., Overton K., Rees K., Nonato L.~G.,
  Bertini E., Silva C.~T.}:
\newblock {SUBPLEX}: A visual analytics approach to understand local model
  explanations at the subpopulation level.
\newblock \emph{IEEE Computer Graphics and Applications 42}, 6 (2022), 24--36.
\newblock \href {https://doi.org/10.1109/MCG.2022.3199727}
  {\path{doi:10.1109/MCG.2022.3199727}}.

\bibitem[YNB21]{Yuan2021An}
\textsc{Yuan J., Nov O., Bertini E.}:
\newblock An exploration and validation of visual factors in understanding
  classification rule sets.
\newblock In \emph{Proceedings of the IEEE Visualization Conference} (2021),
  VIS '21, pp.~6--10.
\newblock \href {https://doi.org/10.1109/VIS49827.2021.9623303}
  {\path{doi:10.1109/VIS49827.2021.9623303}}.

\bibitem[YXX{\etalchar{*}}21]{Yuan2021Evaluation}
\textsc{Yuan J., Xiang S., Xia J., Yu L., Liu S.}:
\newblock Evaluation of sampling methods for scatterplots.
\newblock \emph{IEEE Transactions on Visualization and Computer Graphics 27}, 2
  (2021), 1720--1730.
\newblock \href {https://doi.org/10.1109/TVCG.2020.3030432}
  {\path{doi:10.1109/TVCG.2020.3030432}}.

\bibitem[ZWLC19]{Zhao2019iForest}
\textsc{Zhao X., Wu Y., Lee D.~L., Cui W.}:
\newblock {iForest}: Interpreting random forests via visual analytics.
\newblock \emph{IEEE TVCG 25}, 1 (Jan. 2019), 407--416.
\newblock \href {https://doi.org/10.1109/TVCG.2018.2864475}
  {\path{doi:10.1109/TVCG.2018.2864475}}.

\bibitem[ZYMW19]{Zhang2019Interpreting}
\textsc{Zhang Q., Yang Y., Ma H., Wu Y.}:
\newblock Interpreting {CNNs} via decision trees.
\newblock In \emph{Proceedings of the IEEE/CVF Conference on Computer Vision
  and Pattern Recognition} (June 2019), CVPR '19, IEEE Computer Society,
  pp.~6254--6263.
\newblock \href {https://doi.org/10.1109/CVPR.2019.00642}
  {\path{doi:10.1109/CVPR.2019.00642}}.

\end{thebibliography}

\end{document}